\pdfoutput=1

\documentclass[11pt]{article}

\usepackage[final]{acl}

\usepackage{times}
\usepackage{latexsym}
\usepackage{amssymb}
\usepackage{hyperref}
\usepackage{lipsum}

\usepackage[T1]{fontenc}

\usepackage[utf8]{inputenc}

\usepackage{microtype}

\usepackage{inconsolata}

\usepackage{graphicx}
\usepackage{booktabs}
\usepackage{multirow}
\usepackage{amsmath}
\newcommand{\datasetname}{\textsc{SelfCorSet}}
\newcommand{\methodname}{SCL}

\title{Self-Correction is More than Refinement: \\
A Learning Framework for Visual and Language Reasoning Tasks}

\author{
\bf Jiayi He\textsuperscript{$^{\spadesuit}$}\thanks{These two authors contribute equally to this work.} ~
Hehai Lin\textsuperscript{$^{\diamondsuit}$}\footnotemark[1] ~
Qingyun Wang\textsuperscript{$^{\heartsuit}$} ~
Yi R. (May) Fung\textsuperscript{$^{\diamondsuit}$} ~
Heng Ji\textsuperscript{$^{\heartsuit}$}\\
$^{\heartsuit}$University of Illinois at Urbana-Champaign ~~$^{\spadesuit}$Georgia Institute of Technology \\
$^{\diamondsuit}$The Hong Kong University of Science and Technology \\
\texttt{jhe478@gatech.edu}~~~\texttt{hengji@illinois.edu}
}

\begin{document}
\maketitle
\begin{abstract}
While Vision-Language Models (VLMs) have shown remarkable abilities, they invariably generate flawed responses. 
Self-correction that instructs models to refine their outputs presents a promising solution to this issue.
Previous studies have mainly concentrated on Large Language Models (LLMs), while the self-correction abilities of VLMs, particularly concerning both visual and linguistic information, remain largely unexamined. This study investigates the self-correction capabilities of VLMs during both inference and fine-tuning stages.
We introduce a Self-Correction Learning (SCL) approach that enables VLMs to learn from their self-generated self-correction data through Direct Preference Optimization (DPO) without relying on external feedback, facilitating self-improvement.
Experimental results demonstrate that although VLMs struggle to self-correct effectively during iterative inference without additional fine-tuning and external feedback, they can enhance their performance and avoid previous mistakes through preference fine-tuning when their generated self-correction data are categorized into preferred and disfavored samples.
This study emphasizes that self-correction is not merely a refinement process; rather, it should enhance models' reasoning ability through additional training, enabling them to generate high-quality responses directly without further refinement.\footnote{Code is available at \href{https://github.com/ivy3h/SCL}{https://github.com/ivy3h/SCL}.}
\end{abstract}

\section{Introduction}
Large Language Models (LLMs) have shown exceptional versatility across numerous natural language processing domains~\cite{sun2025personadbefficientlargelanguage,chen2024generalindustrialintelligencesurvey}. 
Building upon the foundational capabilities of LLMs, Vision-Language Models (VLMs)~\cite{liu2024improved,zhu2024minigpt} integrate visual recognition and language understanding by combining pre-trained LLMs and vision models through instruction fine-tuning, leading to significant advancements in multimodal tasks~\cite{huang2024pixelsinsightssurveyautomatic,peng2024data}.

\begin{figure}[t]
    \centering
    \includegraphics[width=1\linewidth]{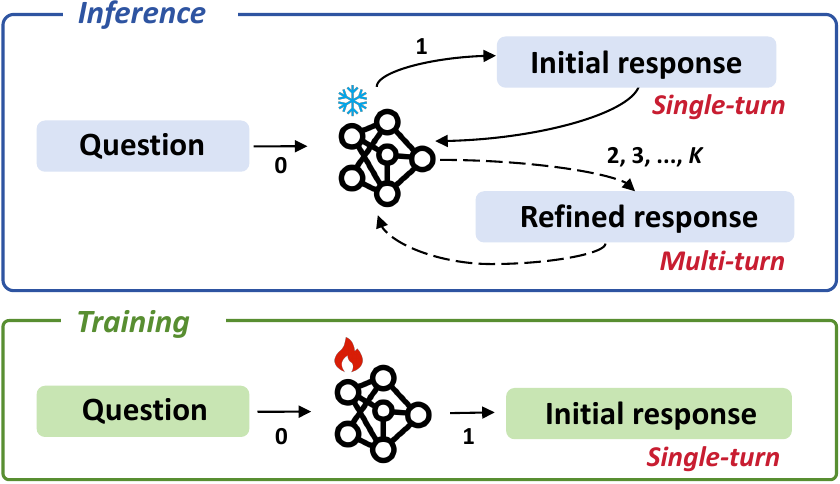}
    \caption{Comparison of inference-based and training-based self-correction. Inference-based methods aim to refine an initial response over $K$ iterations, keeping model parameters fixed. Training-based methods focus on training the model to directly generate high-quality initial responses without iterative refinement.}
    \label{fig:comparison}
\end{figure}

Despite their robust vision-language understanding capabilities, VLMs are still prone to generating inaccurate information~\cite{wu-etal-2024-macaroon,fan2025v2rbenchholisticallyevaluatinglvlm,he2025mmboundaryadvancingmllmknowledge,zhang2025vlm2benchcloserlookvlms}.
Self-correction, an approach enabling models to identify and rectify mistakes in their outputs~\cite{kamoi2024can}, is thus a promising method for enhancing the quality of VLM-generated responses.
While previous studies have predominantly focused on self-correction in LLMs, the self-correction capabilities of VLMs remain underexplored.
Given that VLMs integrate visual and linguistic information during reasoning, self-correction in VLMs presents additional challenges. This complexity arises from the inherent difficulty of accurately aligning and rectifying multimodal data, underscoring the critical need for a systematic investigation into VLM self-correction to advance their performance in vision-language reasoning tasks.

Existing self-correction strategies focus on the inference stage without parameter updates~\cite{madaan2024self,shinn2024reflexion,li2024confidence,liu2024largelanguagemodelsintrinsic}.
These methods typically involve instructing models to revise their initially generated answers by leveraging self-correction prompts. 
While prior work has demonstrated the effectiveness of inference-time self-correction in improving LLM performance on reasoning tasks~\cite{madaan2024self}, and such approaches offer advantages like no dedicated training cost and operational simplicity, recent studies have reported contradictory findings~\cite{huang2023large,xu-etal-2024-pride}.
This controversy highlights two main shortcomings of inference-time self-correction: (1) \textit{Unreliable performance}: The effectiveness of self-correction is highly sensitive to the formulation of self-correction prompts~\cite{li2024confidence}. (2) \textit{Limitations of models' reasoning abilities}: Without additional training to enhance their intrinsic reasoning capabilities, models often struggle to effectively self-correct when confronted with identical challenging tasks~\cite{kamoi2024can}. 

Besides these two challenges, a crucial distinction between existing self-correction methods during the inference stage and the more natural human self-correction process lies in their \textit{correction goals}.
As illustrated in Figure~\ref{fig:comparison}, the former approach focuses on better \textit{refinement}, specifically enabling the model to correct its initial response through additional revisions~\cite{madaan2024self}.
Conversely, the latter approach emphasizes better \textit{initial generation}, aiming to provide the correct answer on the first attempt without subsequent revisions~\cite{zhang2024self,tong2024can}.
This discrepancy indicates that current inference-time self-correction methods offer only a temporary solution for rectifying mistakes. While a model can correct mistakes in its generated content through iterative self-correction, its underlying reasoning ability remains unchanged. Consequently, the model may continue to produce suboptimal answers when faced with identical questions in the future, leading to inefficient resource expenditure on repeated refinement.
Therefore, we emphasize the \textbf{ultimate aim} of self-correction: \textbf{\textit{not merely to fix initial mistakes but to fundamentally improve the model's capability to generate correct answers directly.} }

In this paper, we investigate the self-correction capabilities of VLMs through two research questions (RQs): (1) \textbf{Inference-based self-correction mechanisms:} Can VLMs perform self-correction during inference without external feedback? (2) \textbf{Training-based self-correction mechanisms:} Can VLMs improve their performance by learning from their self-correction process and prevent similar future errors?
Both RQs emphasize the concept of \textit{self}, exploring the intrinsic abilities of VLMs to self-correct autonomously.
For inference-based mechanisms, we design three distinct visual self-correction prompts. These prompts guide VLMs to critically examine their initial responses by scrutinizing input image details, understanding contextual cues, and comprehensively interpreting scenes.
For training-based mechanisms, we propose \textbf{Self-Correction Learning (SCL)}, which utilizes Direct Preference Optimization (DPO)~\cite{rafailov2024direct} to empower VLMs to self-improve by learning from their own generated self-correction preference data. 
The preference dataset, \datasetname{}, is constructed based on the intrinsic self-correction process during inference, where we designate the correct responses as preferred and the incorrect responses as disfavored.

We evaluate the intrinsic self-correction abilities of VLMs and the efficacy of \methodname{} across multiple-choice questions (MCQ) benchmarks.
Experimental results demonstrate that while VLMs initially struggle with intrinsic self-correction, they can benefit from learning from their self-correction samples.
Specifically, VLMs fine-tuned with \methodname{} are better able to avoid previous errors and exhibit superior performance compared to existing preference optimization methods for VLMs. 

Our main contributions are as follows:
\textbf{Firstly}, we define the ultimate objective of self-correction as not only rectifying initial errors but fundamentally enhancing a model's ability to generate accurate responses directly.
\textbf{Secondly}, we systematically evaluate the self-correction abilities of VLMs during inference by developing three visual self-correction prompts. We further analyze the reliability of inference-based self-correction mechanisms.
\textbf{Thirdly}, we introduce Self-Correction Learning (\methodname{}), a novel approach that enables VLMs to self-improve via DPO by leveraging self-generated preference data. Our findings demonstrate the effectiveness of \methodname{} and underscore the advantages of training-based self-correction mechanisms.

\section{Related Work}
\noindent \textbf{Vision-Language Models and Preference Fine-Tuning.} 
Vision-Language Models (VLMs), such as GPT-4o~\cite{gpt-4o}, MiniGPT-4~\cite{zhu2024minigpt}, and LLaVA-1.5~\cite{liu2024improved}, integrate the encoding of visual and textual data to solve various multimodal tasks such as image classification~\cite{peng2024data} and action recognition~\cite{deng2024vg4d}.
Human preference alignment techniques have been widely applied to VLMs to train these models to generate content aligning with human intentions~\cite{chen2024dress}. 
For instance, Silkie constructs a large-scale multimodal preference dataset annotated by GPT-4V, and distills these preferences into VLMs via Direct Preference Optimization (DPO)~\cite{li2023silkiepreferencedistillationlarge}.
Preference Optimization in VLLM with AI-Generated Dispreferences (POVID) utilizes preference fine-tuning to mitigate hallucinations~\cite{zhou2024aligning}.
Calibrated Self-Rewarding (CSR) incorporates an iterative learning and rewarding paradigm into preference fine-tuning for modality alignment~\cite{zhou2024calibrated}.
Similarly, Inner Monologue Multi-Modal Optimization (IMMO) utilizes a combination of supervised learning and reinforcement learning approaches, performing an inner monologue to enhance model performance on complex vision-language tasks~\cite{yang2024tackling}.
While prior studies primarily achieve VLM alignment with human preferences through external feedback from humans or other language models, our work focuses on the self-improvement of VLMs through preference fine-tuning.

\textbf{Intrinsic Self-Correction in Large Language Models.}
Self-correction in Large Language Models (LLMs) aims to guide these models in rectifying their flawed generated content, such as harmful outputs~\cite{helbling2023llm}.
Intrinsic self-correction refers to a self-correction paradigm during inference where a model revises its output solely by leveraging its inherent capabilities and the input context, without external feedback~\cite{huang2023large,kamoi2024can,liu2024largelanguagemodelsintrinsic}.
This iterative, multi-turn self-correction process is distinct from single-turn test-time inference of GPT-o1~\cite{gpt-o1}.
While recent research has demonstrated the effectiveness of intrinsic self-correction~\cite{madaan2024self,shinn2024reflexion,dhuliawala2024chain,li2025mentalarenaselfplaytraininglanguage}, some studies indicate that LLMs face challenges with it. For instance, intrinsic self-correction may sometimes decrease output quality~\cite{huang2023large} and potentially introduce bias~\cite{xu-etal-2024-pride}.
These conflicting results suggest that the self-correction ability of LLMs remains unreliable without external feedback.
Previous work has primarily explored the intrinsic self-correction abilities of LLMs on unimodal tasks like arithmetic reasoning. 
Our study investigates the intrinsic self-correction abilities of VLMs on visual and language reasoning tasks.

\begin{figure*}[t]
    \centering
    \includegraphics[width=1\linewidth]{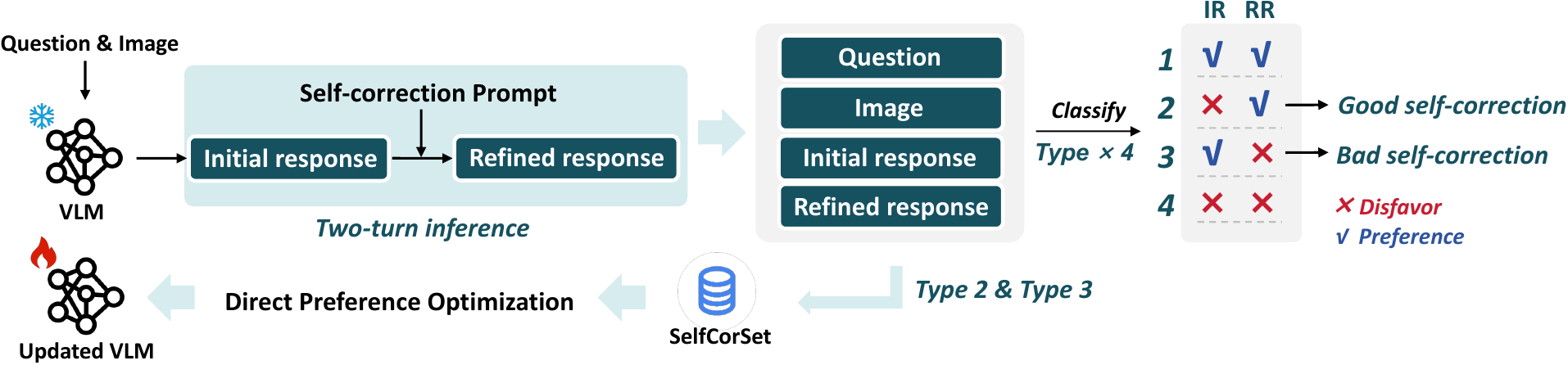}
    \caption{\methodname{} begins with intrinsic self-correction applied to the VLM, generating four types of self-correction samples. Correct responses from Type 2 and incorrect responses from Type 3 samples are designated as preferences and disfavors, respectively, to construct the \datasetname{} preference dataset. The VLM then undergoes DPO on \datasetname{} for self-improvement.}
    \label{fig:framework}
\end{figure*}

\textbf{Improvement in Language Models and Vision-Language Models.}
The enhancement of Language Models (LMs) can be broadly classified into self-improvement and external improvement. 
Self-improvement in LMs relies on their ability to learn from the data they generate, whereas external improvement involves leveraging external models or tools to enable LMs to learn from provided data.
Regarding self-improvement, \citet{huang2022large,wang2025calmunleashingcrosslingualselfaligning} show that LLMs can self-improve by learning from self-generated data selected via self-consistency~\cite{wang2022self}.
\citet{zhang2024self,wu2024aligningllmsindividualpreferences,huang2025mactuningllmmulticompositionalproblem} leverage the LLM's self-evaluation capability to generate training signals, such as for mitigating factual inaccuracies and enhancing response personalization.
\citet{kumar2024traininglanguagemodelsselfcorrect} introduce SCoRe, a multi-turn online reinforcement learning method that achieves self-correction improvements by training on self-generated data and utilizing a two-stage training process to prevent behavior collapse.
Additionally, \citet{wang-etal-2025-enhancing} propose Self-Improvement Modality Alignment (SIMA) which uses in-context self-critic to improve the modality alignment of VLMs. 
In SIMA, the model generates two one-turn responses for each question using greedy decoding and temperature sampling. It is then prompted to critique these responses as preferred or disfavored, thereby constructing a preference dataset.
Distinguished from SIMA, our work constructs a preference dataset using pairs of initial and revised responses from VLMs, generated during intrinsic self-correction.

For external improvement, 
\citet{tong2024can} demonstrate that error data generated by strong LLMs can enhance the reasoning capabilities of weaker LLMs.
\citet{han2024small} show that small LMs can enhance their self-correction capabilities through instruction fine-tuning. However, this method still requires the generation of self-modification responses.
Our work emphasizes that the goal of self-correction is not only to correct mistakes iteratively, but rather to enhance the models' ability to produce correct answers directly. 
We explore whether VLMs can improve the quality of their responses without further refinement steps by leveraging both their successful and erroneous intrinsic self-correction data for self-improvement.

\section{Methodology}
The framework of Self-Correction Learning (SCL) is illustrated in Figure~\ref{fig:framework}, comprising three stages: inference, dataset construction, and preference fine-tuning.
In the inference stage, we propose three visual self-correction prompts and investigate the intrinsic self-correction capabilities of VLMs to address \textbf{RQ1}.
Subsequently, in the dataset construction and fine-tuning stage, we create \datasetname{} for each VLM based on its intrinsic self-correction and explore \textbf{RQ2} through DPO.

\subsection{Inference: Intrinsic Self-Correction}
\label{inference}
Intrinsic self-correction comprises two distinct stages: initial answer generation and refined answer generation.
During the initial answer generation stage, a \textbf{Standard Prompt} (\textbf{SP}) presents the complete question to the VLM, ensuring all requirements are included. For refined answer generation, the VLM engages in an iterative multi-turn process to enhance its initial responses.
Due to computational resource constraints, we limit VLMs to a single refinement turn.
We apply a critical prompt~\cite{huang2023large} and develop three visual self-correction prompts to evaluate VLMs' intrinsic self-correction.
The critical prompt directly guides models to detect issues in their initial responses. The visual self-correction prompts instruct models to identify problems by scrutinizing input image details, comprehending the portrayed context, and comprehensively interpreting entire scenes.
Here are the prompts:

(1)~\textbf{Critical Prompt} (\textbf{CP}): Review your previous answer and find problems with your answer. Based on the problems you found, improve your answer.

(2)~\textbf{Comprehensive Detail Prompt} (\textbf{VP-1}): Review your previous answer and ensure that all relevant aspects of the image have been considered. Are there any elements or details that you missed? Based on your review, improve your answer.

(3)~\textbf{Contextual Understanding Prompt} (\textbf{VP-2}): Review your contextual understanding of the image. Have you correctly interpreted the overall context and purpose of the scene? Based on your review, improve your answer.

(4)~\textbf{Comprehensive Scene Analysis Prompt} (\textbf{VP-3}): Review your answer and ensure that your understanding of the image is comprehensive and detailed. Are there any aspects of the scene that you have omitted or misinterpreted? Based on your review, improve your answer.

\begin{figure*}[t]
    \centering
    \includegraphics[width=1\linewidth]{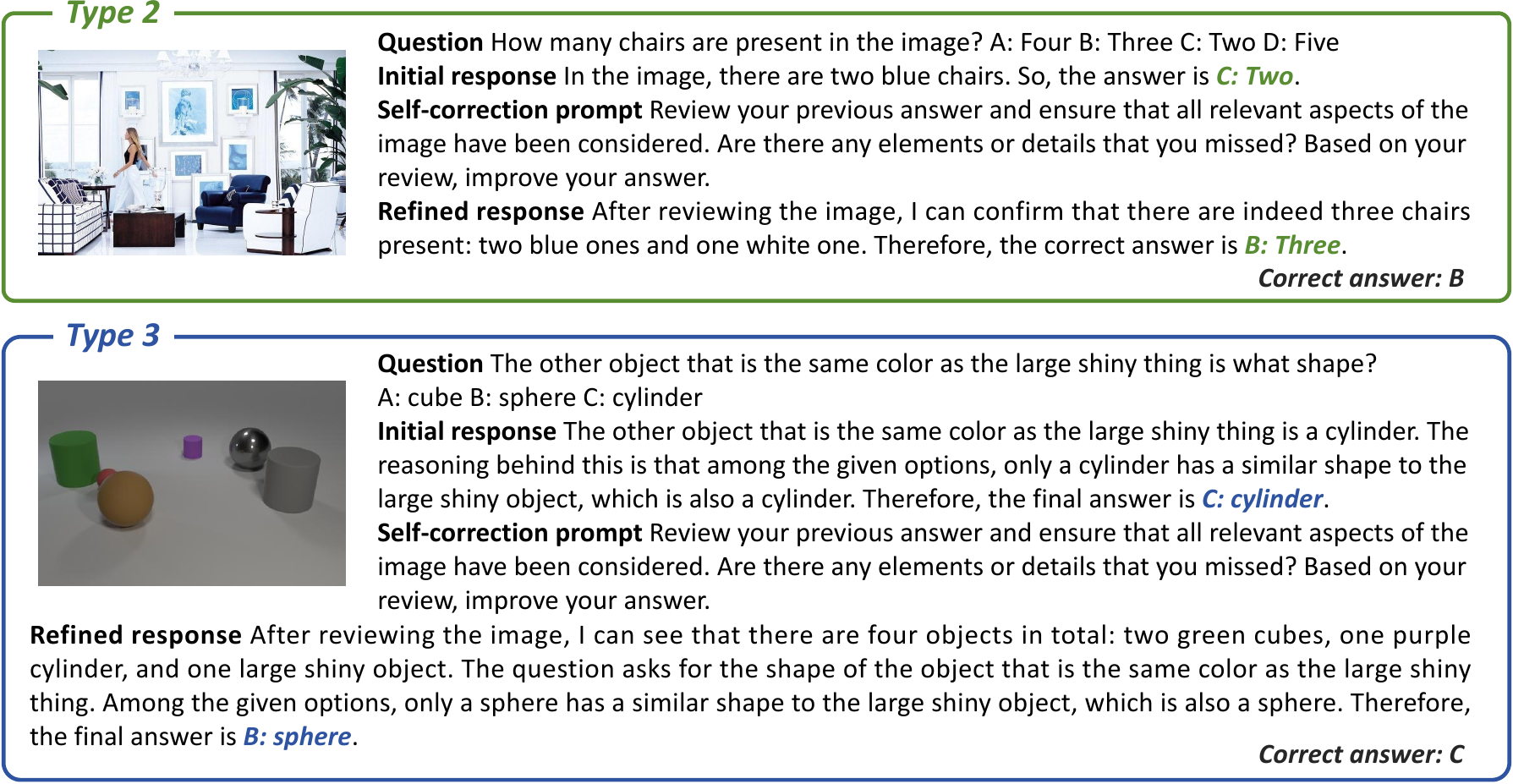}
    \caption{Two examples of intrinsic self-correction processes generated by InternLM-XComposer-2-7B.}
    \label{fig:example}
\end{figure*}

\begin{table*}[t]
\centering
\resizebox{0.96\textwidth}{!}{
\setlength{\tabcolsep}{1.5mm}
\begin{tabular}{ll|cccccccc}
\toprule
& & RealWorldQA & MMStar & MMBench & SEEDBench & ScienceQA & MMT-Bench & Rank\\
\midrule
\multirow{5}{*}{MiniCPM-Llama3-V2.5} & SP & \textbf{61.70} & \textbf{50.40} & \textbf{79.00} & \underline{66.80} & \textbf{75.40} & \textbf{49.00} & \textbf{1.17}\\
& + CP & 38.56 & 40.20 & 68.60  & 62.80 & 69.40 & 37.80 & 4.17\\
& + VP-1 & \underline{48.50} & 46.20 & \underline{76.40} & 64.80 & \underline{73.00} & 37.00 & \underline{3.00}\\
& + VP-2 & 47.32 & \underline{48.40} & 61.00 & 64.40 & 69.00 & 38.00 & 3.33\\
& + VP-3 & 43.00 & 48.00 & 54.00 & \textbf{69.80} & 54.40 & \underline{40.60} & 3.33\\
\midrule 
\multirow{5}{*}{InternLM-XComposer-2-7B} & SP & \textbf{60.13} & \underline{47.40} & \textbf{76.80} & \textbf{69.00} & \textbf{78.20} & \textbf{48.80} & \textbf{1.17}\\
& + CP & 53.86 & 37.00 & 61.60 & 58.00 & 52.60 & 38.40 & 4.83\\
& + VP-1 & 54.50 & \textbf{48.00} & \underline{70.80} & \underline{67.00} & \underline{62.60}
 & 38.60 & \underline{2.50}\\
& + VP-2 & \underline{55.03} & 45.00 & 64.40 & 60.60 & 49.00
 & \underline{41.40} & 3.00\\
& + VP-3 & 54.51 & 39.00 & 61.80 & 59.60 & 58.80
 & 39.00 & 3.50\\
\midrule 
\multirow{5}{*}{LLaVA-V1.5-7B} & SP & \textbf{50.46} & \textbf{32.20} & \textbf{68.40} & \textbf{65.60} & \textbf{65.80} & \textbf{36.00} & \textbf{1.00}\\
& + CP & 36.60 & \underline{24.00} & 54.00 & 36.20 & 56.80 & \underline{32.00} & 2.83\\
& + VP-1 & \underline{43.01} & 22.80 & \underline{57.20} &\underline{42.40} & \underline{58.20} & 29.00 & \underline{2.33}\\
& + VP-2 & 17.78 & 18.60 & 45.40 & 29.00 & 45.80 & 12.00 & 5.00\\
& + VP-3 & 36.21 & 20.40 & 54.00 & 37.00 & 54.80 & 28.40 & 3.67\\ 
\bottomrule
\end{tabular}
}
\caption{Quantitative comparisons (\%) of MiniCPM-Llama3-V-2.5, InternLM-XComposer-2-7B, and LLaVA-V1.5-7B with intrinsic self-correction. Best results are \textbf{bolded}, and second-best are \underline{underlined}. Rank indicates the overall performance ranking across the evaluated benchmarks (lower is better).}
\label{table:intrinsic}
\end{table*}

\subsection{Data Construction: \datasetname{}}
We construct the preference dataset \datasetname{} using intrinsic self-correction outputs generated from multiple-choice question (MCQ) samples. 
These MCQ samples are obtained from several multimodal MCQ datasets, including MMStar~\cite{chen2024we}, MMBench~\cite{10.1007/978-3-031-72658-3_13}, SEEDBench~\cite{li2023seed}, ScienceQA~\cite{lu2022learn}, MMT-Bench~\cite{ying2024mmtbench}, and MMEvalPro~\cite{huang-etal-2025-mmevalpro}.
For MMEvalPro, all available samples are utilized.
For the remaining datasets, 500 samples are randomly selected for evaluation, with the rest allocated for \datasetname{} construction.
Intrinsic self-correction is conducted using the VP-1 prompt, identified as the best-performing self-correction prompt overall (Table~\ref{table:intrinsic}).

The construction process for the \datasetname{} dataset, using a VLM denoted as $M$, is outlined as follows: Given an MCQ sample $s=\left \{ Question, Image, Answer \right \}$, we first perform intrinsic self-correction with VLM $M$ to obtain a self-correction sample $s'=\left \{ Question, Image, IR, RR \right \}$, where $IR$ denotes the initial response and $RR$ is the refined response.
In this process, $M$ generates $IR$ when prompted with the initial question and image. Subsequently, $RR$ is generated as a revision of $IR$ after $M$ receives the self-correction prompt.
Based on the correctness of $IR$ and $RR$, each self-correction sample $s'$ is classified into four distinct types:
Specifically, the corresponding correctness of $IR$ and $RR$ for the four types of samples are as follows:
\textbf{Type 1} (correct$\Rightarrow$correct), \textbf{Type 2} (incorrect$\Rightarrow$correct), \textbf{Type 3} (correct$\Rightarrow$incorrect), and \textbf{Type 4} (incorrect$\Rightarrow$incorrect). 
The $\Rightarrow$ symbol denotes the transition of correctness from the initial response ($IR$) to the refined response ($RR$).
Type 2 samples represent successful self-corrections, as the model effectively revises an incorrect $IR$ into a correct $RR$. Conversely, Type 3 samples indicate detrimental self-corrections, where the model erroneously alters a correct $IR$ to an incorrect one.
The \datasetname{} preference dataset is then constructed using Type 2 and Type 3 samples according to the following criterion: $RR$ from Type 2 samples and $IR$ from Type 3 samples are designated as preferred responses, as they represent correct answers. 
In contrast, $IR$ from Type 2 samples and $RR$ from Type 3 samples are classified as disfavored responses.
Figure~\ref{fig:example} presents two examples of Type 2 and Type 3 self-correction outputs generated by InternLM-XComposer-2-7B. 
In the Type 2 example, the model successfully revises an incorrect $IR$ (C: Two) to a correct $RR$ (B: Three) after reviewing the image. Conversely, the Type 3 example shows a detrimental self-correction, where the model incorrectly alters an initially correct $IR$ (C: cylinder) to an incorrect $RR$ (B: sphere).

We construct three distinct \datasetname{}, one for each of the three evaluated VLMs: LLaVA-V1.5-7B~\cite{liu2024improved}, LLaVA-V1.5-13B~\cite{liu2024improved}, and MiniCPM-Llama3-V2.5~\cite{yao2024minicpm}. 
This construction emphasizes the uniqueness of \textit{self}, as different VLMs possess distinct intrinsic self-correction behaviors and thus generate unique sets of self-correction samples.
Each sample comprises a question, an image, a preferred response, and a disfavored response.
Data samples are provided in Appendix~\ref{Data Examples}.

\subsection{Training: Learn from Self-Correction}
We fine-tune the VLM using Direct Preference Optimization (DPO)~\cite{rafailov2024direct} on its dedicated preference dataset, \datasetname{}.
We denote \datasetname{} as $\mathcal{D}_{sc}=\{(Q^{(i)},I^{(i)},R_{c}^{(i)},R_{r}^{(i)})\}_{i=1}^{N}$, where $Q^{(i)}$ represents the input question, $I^{(i)}$ is the corresponding image, $R_{c}^{(i)}$ is the preferred response, and $R_{r}^{(i)}$ is the disfavored response.
The DPO loss is defined as:
\begin{equation}
\resizebox{0.85\hsize}{!}{$
\mathcal{L}(\pi_{\theta}; \pi_{\text{ref}}) = -\mathbb{E}_{(Q, I, R_{c}, R_{r})}\left [ \log \sigma f(\pi_{\theta}; \pi_{\text{ref}})  \right ]
$}
\end{equation}
\begin{equation}
\resizebox{0.89\hsize}{!}{$
f(\pi_{\theta}; \pi_{\text{ref}}) = \beta \left ( \log\frac{\pi_{\theta}(R_{c}|Q,I)}{\pi_{\text{ref} }(R_{c}|Q,I)} 
- \log\frac{\pi_{\theta}(R_{r}|Q,I)}{\pi_{\text{ref} }(R_{r}|Q,I)} \right )
$}
\end{equation}
where $\sigma$ represents the logistic sigmoid function, $\pi _{\theta }$ denotes the current VLM policy, $\pi _{\text{ref}}$ denotes the reference policy, and $\beta$ is a hyperparameter that controls the strength of the penalty against deviating from the reference policy.
Both $\pi _{\theta }$ and $\pi _{\text{ref}}$ are initialized with the same weight.

\section{Experiments}
\subsection{Experimental Settings}
\textbf{Test Models.} 
For intrinsic self-correction evaluation, we conduct experiments on three open-source VLMs: MiniCPM-Llama3-V2.5~\cite{yao2024minicpm}, InternLM-XComposer-2-7B~\cite{internlmxcomposer2}, and LLaVA-V1.5-7B~\cite{liu2024improved}. 
MiniCPM-Llama3-V2.5 is an advanced VLM with a total of 8B parameters.
InternLM-XComposer-2-7B is designed for the comprehension and composition of free-form text-image pairs. 
LLaVA-V1.5-7B is a widely used VLM trained with visual instructions.
For self-correction training evaluation, we conduct experiments on MiniCPM-Llama3-V2.5, InternLM-XComposer-2-7B, LLaVA-V1.5-7B, and LLaVA-V1.5-13B.

\textbf{Evaluation Benchmarks.} 
We conduct evaluations on eight multimodal multiple-choice question (MCQ) benchmarks: RealWorldQA~\cite{RealWorldQA}, MMStar~\cite{chen2024we}, MMBench-en~\cite{10.1007/978-3-031-72658-3_13}, SEEDBench~\cite{li2023seed}, ScienceQA~\cite{lu2022learn}, MMT-Bench~\cite{ying2024mmtbench}, MMMU~\cite{yue2023mmmu}, and AI2D~\cite{kembhavi2016diagram}. For intrinsic self-correction, we use the first six datasets with the following sample counts for evaluation: RealWorldQA (765), MMStar (500), MMBench (500), SEEDBench (500), ScienceQA (500), MMT-Bench (500). 
We further incorporate two benchmarks to evaluate fine-tuned models: MMMU (1050) and AI2D (3088).
We adopt accuracy and average rank as the metrics.

\textbf{Training Baselines.} We compare \methodname{} with three preference optimization methods: POVID~\cite{zhou2024aligning}, CSR~\cite{zhou2024calibrated}, and SIMA~\cite{wang-etal-2025-enhancing}.
POVID leverages GPT to enhance the quality of ground truth answers and employs DPO for training.
CSR incorporates iterative learning and a reward-based paradigm into its preference fine-tuning process.
SIMA deploys in-context self-critic to construct its preference dataset and utilizes DPO to enhance the comprehension capabilities of VLM.
We also compare \methodname{} with Supervised Fine-Tuning (SFT), which directly utilizes the preferred responses of \datasetname{} for fine-tuning.

\textbf{Implementation Details.} Intrinsic self-correction is conducted on a total of 26,981 samples. The resulting \datasetname{} dataset sizes for each VLM are as follows: MiniCPM-Llama3-V2.5 (1,853), InternLM-XComposer-2-7B (2,361), LLaVA-V1.5-7B (4,797), and LLaVA-V1.5-13B (738).
For computational efficiency, we adopt LoRA\cite{hu2022lora}, a widely recognized Parameter-Efficient Fine-Tuning technique, for training.  
We set the LoRA rank as 8.
Training for 7/8B models is conducted on a single 4090 24GB GPU with 1.5 GPU hours per epoch.
Training for 13B models is conducted on a single V100 32GB GPU with 1.5 GPU hours for three epochs.

\begin{table*}[t]
\centering
\setlength{\tabcolsep}{1mm}
\resizebox{\textwidth}{!}{
\begin{tabular}{l|c c c c c c c c c}
\toprule
& RealWorldQA & MMStar & MMBench & SEEDBench & ScienceQA & MMT-Bench & MMMU & AI2D & Rank\\
\midrule 
LLaVA-V1.5-7B&50.90&32.97&70.24&66.64&65.48&35.40&32.25&52.71&4.50\\
+POVID&51.50&33.68&\textbf{71.44}&65.52&64.60&35.04&33.96&\underline{53.59}&3.25\\
+CSR&51.03&32.59&70.44&65.12&64.76&33.76&\underline{34.63}&53.34&4.25\\
+SIMA&49.41&32.40&\underline{71.04}&64.68&64.68&34.84&\textbf{35.14}&53.01&4.38\\
+SFT&\underline{52.21}&\underline{35.59}&70.72&\underline{67.12}&\underline{66.48}&\underline{36.36}&32.95&52.73&\underline{3.00}\\
+\textbf{\methodname (Ours)}&\textbf{52.71}&\textbf{36.11}&71.00&\textbf{67.84}&\textbf{66.56}&\textbf{36.96}&33.62&\textbf{54.81}&\textbf{1.63}\\\midrule
LLaVA-V1.5-13B&54.85&35.25&74.40&\textbf{69.40}&71.36&40.08&34.88&55.95&2.75\\
+SFT&\textbf{56.59}&\underline{36.33}&\underline{75.16}&\underline{69.32}&\underline{71.88}&\underline{40.48}&\underline{35.29}&\underline{56.00}&\underline{1.88}\\
+\textbf{\methodname (Ours)}&\underline{56.03}&\textbf{37.60}&\textbf{75.40}&68.56&\textbf{72.16}&\textbf{41.16}&\textbf{35.87}&\textbf{58.92}&\textbf{1.38}\\\midrule
InternLM-XComposer-2-7B & 60.45 & 47.70 & 76.60 & 68.90 & 78.30 & \underline{49.20} & 56.48 & 80.27 & 3.00\\
+SFT & \underline{62.03} & \underline{48.90} & \textbf{77.80} & \underline{70.20} & \underline{79.50} & \textbf{50.20} & \underline{57.95} & \underline{80.66} & \underline{1.75}\\
\textbf{+SCL (Ours)} & \textbf{62.23} & \textbf{49.60} & \underline{77.00} & \textbf{70.60} & \textbf{79.90} & \textbf{50.20} & \textbf{58.20} & \textbf{81.11} & \textbf{1.13}\\
\midrule
MiniCPM-Llama3-V2.5  & 61.70 & 50.40 & 79.00 & 66.80 & 75.40 & 49.00 & 45.24 & 77.56 & 3.00\\
+SFT& \underline{62.35} & \underline{52.40} & \underline{80.80} & \underline{68.40} & \underline{76.00} & \underline{49.80} & \underline{47.43} & \underline{78.01} & \underline{2.00}\\
+\textbf{\methodname (Ours)} & \textbf{63.53} & \textbf{53.00} & \textbf{81.40} & \textbf{69.20} & \textbf{76.40} & \textbf{50.40} & \textbf{47.52} & \textbf{78.72} & \textbf{1.00}\\
\bottomrule
\end{tabular}
}
\caption{Quantitative comparisons (\%) of LLaVA-V1.5-7B, LLaVA-V1.5-13B, InternLM-XComposer-2-7B, and MiniCPM-Llama3-V2.5 with \methodname{} and baseline methods. Best results are \textbf{bolded}, and second-best are \underline{underlined}. Rank indicates the overall performance ranking across the evaluated benchmarks (lower is better).}
\label{table:finetune}
\end{table*}

\subsection{Results and Analysis}
\begin{figure}[t]
    \centering
    \includegraphics[width=1\linewidth]{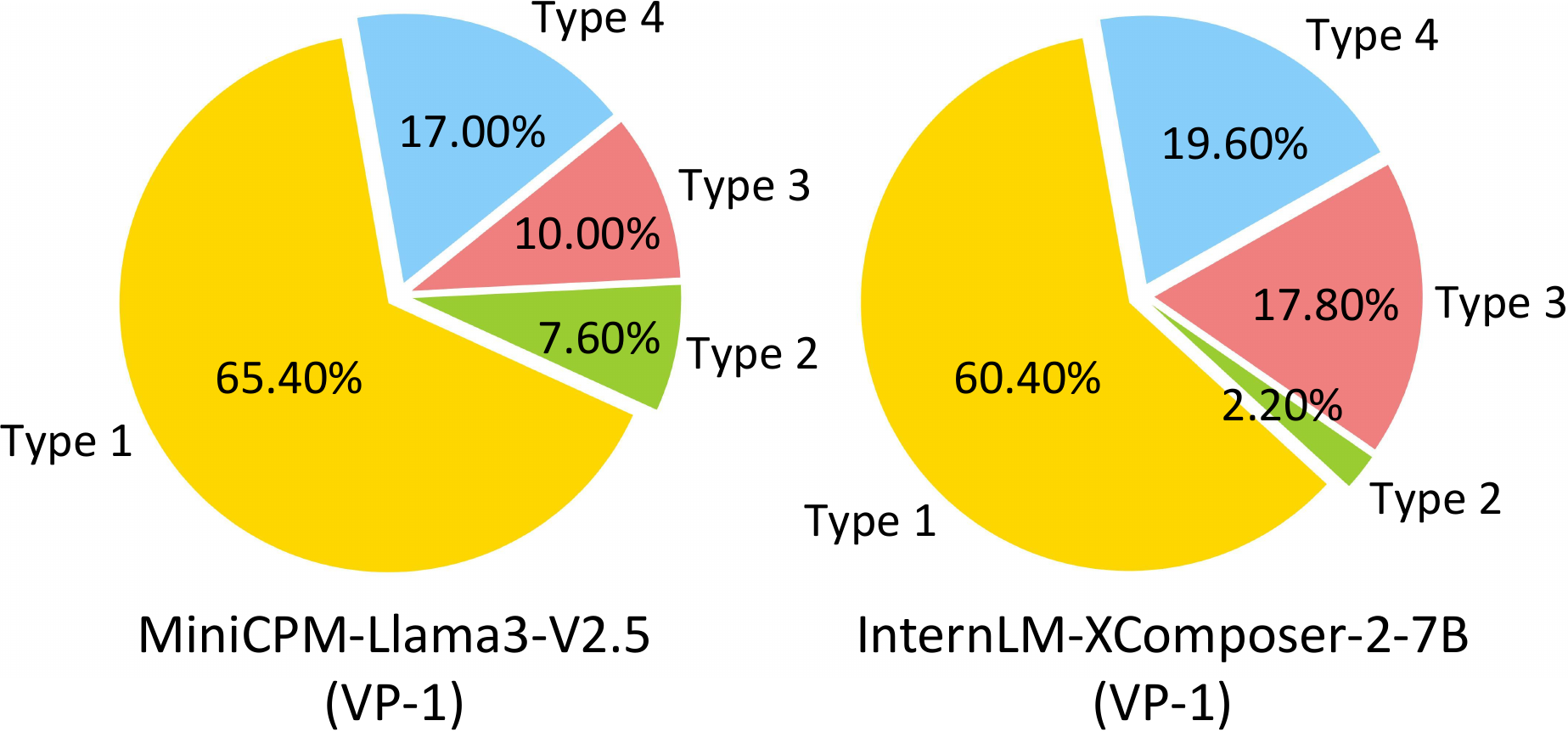}
    \caption{Distribution of self-correction examples of MiniCPM-Llama3-V2.5 and InternLM-XComposer-2-7B under VP-1 on ScienceQA.}
    \label{fig:distribution}
\end{figure}

\textbf{VLMs struggle in intrinsic self-correction.}
Table~\ref{table:intrinsic} presents the results of VLMs regarding intrinsic self-correction. 
It can be observed that, in the majority of cases, VLMs achieve higher accuracy prior to intrinsic self-correction. For instance, MiniCPM-Llama3-V2.5's accuracy is 61.70\% in its initial generation but drops to only 48.50\% after intrinsic self-correction under the VP-1 prompt. These results consistently indicate that VLMs struggle to achieve effective performance improvement through intrinsic self-correction. Furthermore, the effectiveness of intrinsic self-correction varies significantly across different models, tasks, and self-correction prompts. While MiniCPM-Llama3-V2.5 improves its performance on SEEDBench with the VP-3 prompt (66.80\% $\rightarrow$ 69.80\%), it fails to self-correct on other benchmarks, highlighting the inherent instability of intrinsic self-correction.

To further illustrate the correctness transition during intrinsic self-correction, Figure~\ref{fig:distribution} displays the distribution of self-correction sample types for MiniCPM-Llama3-V2.5 and InternLM-XComposer-2-7B on ScienceQA under the VP-1 prompt.
The proportion of Type 3 samples (correct$\Rightarrow$incorrect) exceeds that of Type 2 (incorrect$\Rightarrow$correct) for both models, indicating that while models can revise incorrect answers, they more frequently erroneously alter initially correct answers. 
Consistent with the findings of~\citet{huang2023large} regarding LLMs, which reports degraded performance following intrinsic self-correction, our results suggest that VLMs struggle to rectify their own answers reliably. 

\textbf{VLMs self-improve from their self-correction examples.} Table~\ref{table:finetune} shows the performance of \methodname{} in comparison to various preference optimization baselines.
\methodname{} achieves state-of-the-art performance for LLaVA-V1.5-7B on six benchmarks. 
While SCL exhibits suboptimal performance on MMMU, these results indicate that the preference data derived from self-correction samples effectively fine-tunes these models.
All four evaluated VLMs achieve the best overall performance as indicated by the Rank metric.
Despite using a relatively small fine-tuning dataset, our findings demonstrate that VLMs can benefit from both successful and erroneous self-correction samples. 
This suggests that VLMs possess a human-like behavior, i.e., effectively learning by internalizing the distinction between successful corrections and the insights gained from analyzing erroneous attempts.
This ability enables them to efficiently enhance their comprehensive reasoning capabilities.

\begin{figure*}[t]
    \centering
    \includegraphics[width=1\linewidth]{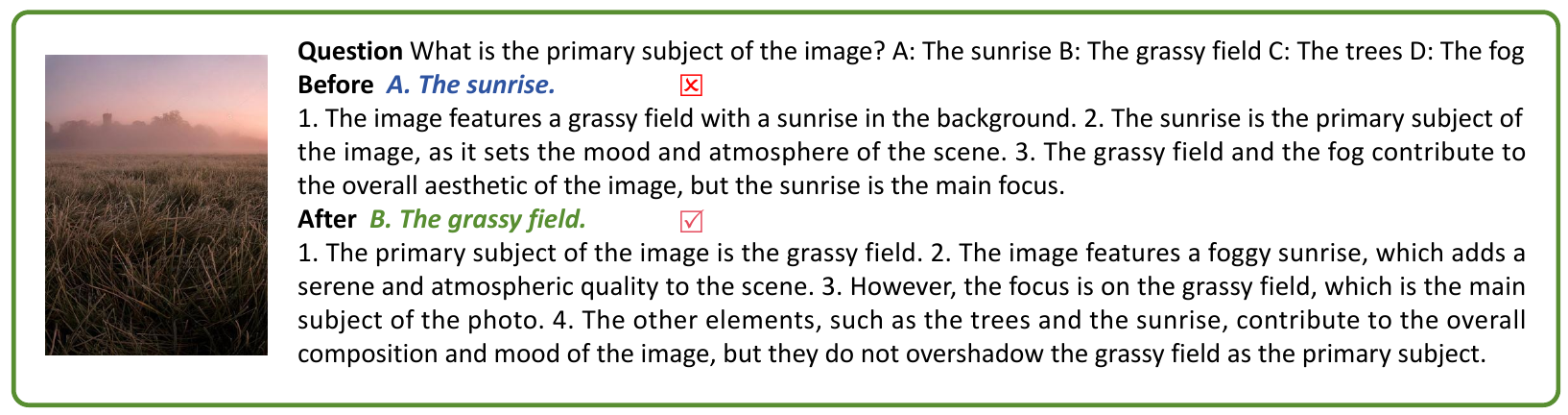}
    \caption{LLaVA-V1.5-7B successfully answers the question after learning from its self-correction samples.}
    \label{fig:case_study}
\end{figure*}

\subsection{Case Study}
Figure~\ref{fig:case_study} presents the initial responses generated by LLaVA-V1.5-7B before and after preference fine-tuning. Prior to fine-tuning, the model incorrectly identified the sunrise as the primary object, misinterpreting it as the element that establishes the mood and atmosphere of the scene. Following fine-tuning with \methodname{}, the model correctly recognizes that the question seeks the object occupying the most spatial area in the image and subsequently produces the correct answer, demonstrating the model's ability to avoid previous errors through preference fine-tuning.

\section{Further Studies and Analysis}
\subsection{Reliability of Successful Intrinsic Self-Correction}
We observe that some successful refinements, classified as Type 2 self-correction cases, stem from the model's incidental guessing of the correct answer after receiving a self-correction prompt, rather than from proper reasoning of the task. For instance, when presented with an image that does not contain a teapot, models might respond with ``The teapot may exist behind the woman...'' and then conclude that a teapot is present in the image after self-correction. This type of refinement reflects a degree of uncertainty.
Moreover, the reasoning behind seemingly successful refinements may not be entirely accurate. For example, in the Type 2 case presented in Figure~\ref{fig:example}, the model correctly identifies the number of chairs after self-correction but fails to accurately determine their colors. 
These observations on the reliability of successful refinements collectively indicate that VLMs possess limited capabilities for truly accurate refinement. Further supporting data is provided in Appendix~\ref{Examples Supporting Further Studies}.

Future research could explore the internal information flow that mediates the transition from initial to refined responses during intrinsic self-correction, for instance, by visualizing attention weights to improve both interpretability and reliability.
More importantly, cases where the final answer is correct but the reasoning is flawed can negatively impact the self-correction mechanism. Future work should focus on designing more detailed evaluation criteria that account for the correctness of intermediate reasoning steps, such as by leveraging a strong model as an MLLM-as-a-judge~\cite{chen2024mllmasajudge}.

\begin{table}[t]
    \centering
    \setlength{\tabcolsep}{1.3mm}
    \resizebox{\columnwidth}{!}{%
    \begin{tabular}{c|cccc}
    \toprule
        & Turn 0 & Turn 1 & Turn 2 & Turn 3 \\
        \midrule
        RealWorldQA (VP-1) & \textbf{61.70} & \underline{48.50} & 39.22& 42.61 \\
       MMStar (VP-2) & \textbf{50.40} & 48.40 & \underline{49.20} & 46.80\\
        MMBench (VP-1) & \textbf{79.00} & \underline{76.40} & 75.60 & 72.60 \\
       SEEDBench (VP-3) & \underline{66.80} & \textbf{69.80} & 63.80 & 64.00\\ 
       \bottomrule
    \end{tabular}
    }
    \caption{Results (\%) of MiniCPM-Llama3-V2.5 with multi-turn intrinsic self-correction.}
    \label{tab:multi-turn}
    \vspace{-1em}
\end{table}

\subsection{Multi-Turn Intrinsic Self-Correction}
Table \ref{tab:multi-turn} presents the results of multi-turn intrinsic self-correction for MiniCPM-Llama3-V2.5 on four benchmarks. Turn 0 represents the initial generation, while Turns 1 to 3 illustrate subsequent intrinsic self-correction iterations. Notably, the accuracy of the refined answers consistently degrades over multiple turns, ultimately falling below the initial accuracy after three correction iterations.
This decline suggests that VLMs find it challenging to achieve effective intrinsic self-correction solely by increasing the number of correction iterations.

Despite this observed degradation, multi-turn intrinsic self-correction may generate a richer, more diverse set of self-correction samples. Each iterative attempt may produce different types of errors when responding to the same question, thereby providing varied erroneous data for preference fine-tuning. Therefore, future research could investigate whether VLMs can derive greater benefits from these additional, potentially more informative samples, even if the multi-turn accuracy diminishes.

\subsection{Failure of VLM Intrinsic Self-Correction}
\vspace{-0.2em}
We attribute the observed failures of intrinsic self-correction in VLMs to several potential factors:
(1) \textbf{Limited ability to judge correctness}~\cite{huang2023large}: VLMs may lack the fundamental reasoning capabilities required to reliably detect and revise their own errors. The inherent multimodal nature of these models—integrating both visual and textual inputs—introduces additional complexity, making it difficult for them to synthesize information across modalities and accurately evaluate correctness.
(2) \textbf{Susceptibility to prompt bias}~\cite{li2024confidence}: As demonstrated in Table\ref{table:intrinsic}, self-correction performance varies notably with different prompts, suggesting a sensitivity to prompt design. Existing self-correction prompts may not sufficiently capture the intricate interplay between visual and textual content, which may impair the models' ability to effectively revise their responses.
(3) \textbf{Lack of a grounding mechanism}: In contrast to self-correction methods that utilize external feedback or oracle labels~\cite{huang2023large,shinn2024reflexion}, intrinsic self-correction operates without explicit supervision. This absence of grounding guidance limits the model's ability to converge on more accurate responses, resulting in ineffective or even detrimental revisions.
Given these challenges, future work could explore the design of more diverse and task-specific self-correction prompts tailored for multimodal settings. Additionally, incorporating a strong teacher model to provide instructional feedback, or integrating oracle labels during the self-correction process, may help improve the reliability of revisions.

\subsection{Effect of the Number of Training Samples}
\begin{figure}[t]
    \centering
    \includegraphics[width=1\linewidth]{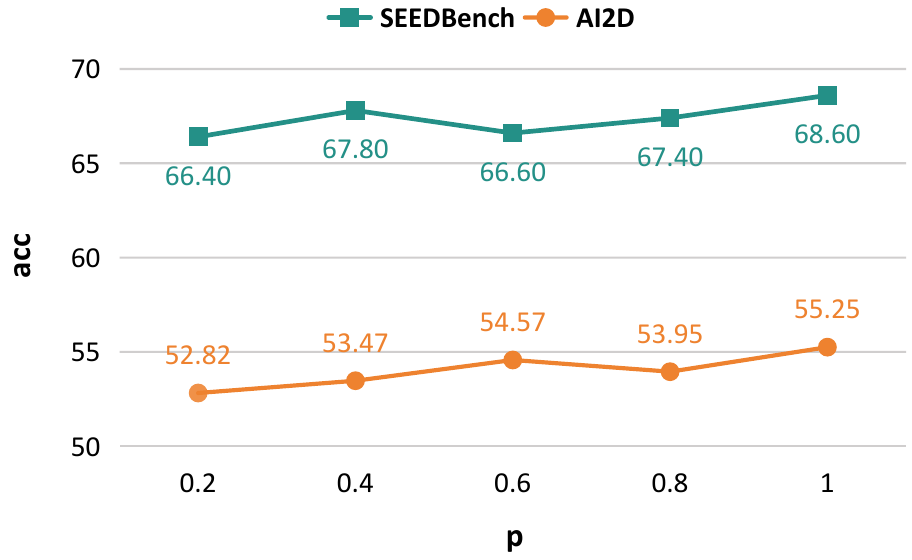}
    \caption{Results of LLaVA-V1.5-7B under different proportions ($p$) of \datasetname{}.}
    \label{fig:subset}
\end{figure}

\vspace{-0.2em}
The limited proportion of Type 2 and Type 3 data, as illustrated in Figure~\ref{fig:distribution}, coupled with the tendency of more advanced models to generate fewer samples of these types, results in a relatively small sample size for \datasetname{}. 
To explore the influence of fine-tuning data quantity, we randomly divide \datasetname{} into five subsets, increasing in 20\% increments from 0\%.
We evaluate the impact of these varying training set sizes on the performance of LLaVA-V1.5-7B on SEEDBench and AI2D, with the accuracy trend illustrated in Figure~\ref{fig:subset}.
Notably, even with smaller training datasets, the fine-tuned model exhibits significant performance gains. For instance, the model fine-tuned on the $p=0.4$ subset achieves an accuracy of 67.80\% on SEEDBench, reflecting a 2.2\% improvement over the untrained model. 
These results indicate that \methodname{} effectively improves the model's performance despite the relatively modest size of \datasetname{}. Moreover, as the number of training samples increases, the overall accuracy consistently improves, showcasing the potential of \methodname{} with larger preference datasets.
Therefore, future research could explore more straightforward and computationally efficient data augmentation methods to scale \datasetname{} and investigate SCL's performance under large-scale data conditions.

\section{Conclusion}
\vspace{-0.3em}
This work investigates the intrinsic self-correction capabilities and limitations of Vision-Language Models (VLMs) during the inference stage, and their potential for self-improvement through fine-tuning, across various multiple-choice question (MCQ) benchmarks.
We propose Self-Correction Learning (SCL), a novel framework that employs Direct Preference Optimization (DPO) to enable VLMs to learn from their own self-correction responses. This approach aims to fundamentally enhance the models' ability to generate correct answers directly, rather than merely refining initial errors.
Our experiments reveal that while VLMs face significant challenges and exhibit unreliable performance during intrinsic self-correction, they can effectively leverage their own self-correction samples (both successful and erroneous) to self-improve without requiring external feedback.

\section*{Limitations}
Given the challenges of evaluating absolute correctness in open-ended multimodal tasks, this study primarily focuses on MCQ benchmarks to assess the accuracy of both initial and refined responses. 
However, this design inherently limits the exploration of more complex vision understanding tasks, such as visual question answering~\cite{chen2025faultgptindustrialfaultdiagnosis}, video question answering~\cite{chen2024autoeval,fu2024video}, and complex transportation system navigation~\cite{li2023understandingdriverpedestrianinteractionspredict}. Future work could investigate fine-grained evaluation methods applicable across a wider range of multimodal tasks, as well as computational resource constraint-aware adaptive reasoning \cite{huang2025adactrladaptivecontrollablereasoning}.

The current construction of \datasetname{} is tailored to each VLM, leveraging the model's own preferred and disfavored responses to enable more targeted self-improvement. However, this design limits generalizability across different models. A potential direction for future work is to develop a unified version of \datasetname{} that can be applied across multiple VLMs. This would involve identifying common error patterns shared among VLMs and designing universal prompts that generalize well across different architectures.

\bibliography{custom}

\begin{thebibliography}{57}
\providecommand{\natexlab}[1]{#1}

\bibitem[{Chen et~al.(2024{\natexlab{a}})Chen, Chen, Zhang, Wang, Liu, Zhou, Zhang, Wan, Zhou, and Sun}]{chen2024mllmasajudge}
Dongping Chen, Ruoxi Chen, Shilin Zhang, Yaochen Wang, Yinuo Liu, Huichi Zhou, Qihui Zhang, Yao Wan, Pan Zhou, and Lichao Sun. 2024{\natexlab{a}}.
\newblock \href {https://openreview.net/forum?id=dbFEFHAD79} {{MLLM}-as-a-judge: Assessing multimodal {LLM}-as-a-judge with vision-language benchmark}.
\newblock In \emph{Forty-first International Conference on Machine Learning}.

\bibitem[{Chen et~al.(2024{\natexlab{b}})Chen, He, Chen, Lv, Tang, Li, Liu, Yang, and Han}]{chen2024generalindustrialintelligencesurvey}
Jiao Chen, Jiayi He, Fangfang Chen, Zuohong Lv, Jianhua Tang, Weihua Li, Zuozhu Liu, Howard~H. Yang, and Guangjie Han. 2024{\natexlab{b}}.
\newblock \href {https://arxiv.org/abs/2409.01207} {Towards general industrial intelligence: A survey of continual large models in industrial iot}.
\newblock \emph{Preprint}, arXiv:2409.01207.

\bibitem[{Chen et~al.(2025)Chen, Huang, Lv, Tang, and Li}]{chen2025faultgptindustrialfaultdiagnosis}
Jiao Chen, Ruyi Huang, Zuohong Lv, Jianhua Tang, and Weihua Li. 2025.
\newblock \href {https://arxiv.org/abs/2502.15481} {Faultgpt: Industrial fault diagnosis question answering system by vision language models}.
\newblock \emph{Preprint}, arXiv:2502.15481.

\bibitem[{Chen et~al.(2024{\natexlab{c}})Chen, Li, Dong, Zhang, Zang, Chen, Duan, Wang, Qiao, Lin, and Zhao}]{chen2024we}
Lin Chen, Jinsong Li, Xiaoyi Dong, Pan Zhang, Yuhang Zang, Zehui Chen, Haodong Duan, Jiaqi Wang, Yu~Qiao, Dahua Lin, and Feng Zhao. 2024{\natexlab{c}}.
\newblock \href {https://arxiv.org/abs/2403.20330} {Are we on the right way for evaluating large vision-language models?}
\newblock \emph{Preprint}, arXiv:2403.20330.

\bibitem[{Chen et~al.(2024{\natexlab{d}})Chen, Lin, Zhang, and Huang}]{chen2024autoeval}
Xiuyuan Chen, Yuan Lin, Yuchen Zhang, and Weiran Huang. 2024{\natexlab{d}}.
\newblock Autoeval-video: An automatic benchmark for assessing large vision language models in open-ended video question answering.
\newblock In \emph{European Conference on Computer Vision}, pages 179--195. Springer.

\bibitem[{Chen et~al.(2024{\natexlab{e}})Chen, Sikka, Cogswell, Ji, and Divakaran}]{chen2024dress}
Yangyi Chen, Karan Sikka, Michael Cogswell, Heng Ji, and Ajay Divakaran. 2024{\natexlab{e}}.
\newblock Dress: Instructing large vision-language models to align and interact with humans via natural language feedback.
\newblock In \emph{Proceedings of the IEEE/CVF Conference on Computer Vision and Pattern Recognition (CVPR)}, pages 14239--14250.

\bibitem[{Deng et~al.(2024)Deng, Li, Li, Tong, Zhao, and Liu}]{deng2024vg4d}
Zhichao Deng, Xiangtai Li, Xia Li, Yunhai Tong, Shen Zhao, and Mengyuan Liu. 2024.
\newblock \href {https://doi.org/10.1109/ICRA57147.2024.10610217} {Vg4d: Vision-language model goes 4d video recognition}.
\newblock In \emph{2024 IEEE International Conference on Robotics and Automation (ICRA)}, pages 5014--5020.

\bibitem[{Dhuliawala et~al.(2024)Dhuliawala, Komeili, Xu, Raileanu, Li, Celikyilmaz, and Weston}]{dhuliawala2024chain}
Shehzaad Dhuliawala, Mojtaba Komeili, Jing Xu, Roberta Raileanu, Xian Li, Asli Celikyilmaz, and Jason Weston. 2024.
\newblock \href {https://aclanthology.org/2024.findings-acl.212} {Chain-of-verification reduces hallucination in large language models}.
\newblock In \emph{Findings of the Association for Computational Linguistics ACL 2024}, pages 3563--3578, Bangkok, Thailand and virtual meeting. Association for Computational Linguistics.

\bibitem[{Dong et~al.(2024)Dong, Zhang, Zang, Cao, Wang, Ouyang, Wei, Zhang, Duan, Cao, Zhang, Li, Yan, Gao, Zhang, Li, Li, Chen, He, Zhang, Qiao, Lin, and Wang}]{internlmxcomposer2}
Xiaoyi Dong, Pan Zhang, Yuhang Zang, Yuhang Cao, Bin Wang, Linke Ouyang, Xilin Wei, Songyang Zhang, Haodong Duan, Maosong Cao, Wenwei Zhang, Yining Li, Hang Yan, Yang Gao, Xinyue Zhang, Wei Li, Jingwen Li, Kai Chen, Conghui He, and 4 others. 2024.
\newblock \href {https://arxiv.org/abs/2401.16420} {Internlm-xcomposer2: Mastering free-form text-image composition and comprehension in vision-language large model}.
\newblock \emph{Preprint}, arXiv:2401.16420.

\bibitem[{Fan et~al.(2025)Fan, Wang, Polisetty, and Fung}]{fan2025v2rbenchholisticallyevaluatinglvlm}
Zhiyuan Fan, Yumeng Wang, Sandeep Polisetty, and Yi~R. Fung. 2025.
\newblock \href {https://arxiv.org/abs/2504.16727} {Unveiling the lack of lvlm robustness to fundamental visual variations: Why and path forward}.
\newblock \emph{Preprint}, arXiv:2504.16727.

\bibitem[{Fu et~al.(2024)Fu, Dai, Luo, Li, Ren, Zhang, Wang, Zhou, Shen, Zhang et~al.}]{fu2024video}
Chaoyou Fu, Yuhan Dai, Yongdong Luo, Lei Li, Shuhuai Ren, Renrui Zhang, Zihan Wang, Chenyu Zhou, Yunhang Shen, Mengdan Zhang, and 1 others. 2024.
\newblock Video-mme: The first-ever comprehensive evaluation benchmark of multi-modal llms in video analysis.
\newblock \emph{arXiv preprint arXiv:2405.21075}.

\bibitem[{Han et~al.(2024)Han, Liang, Shi, He, and Xiao}]{han2024small}
Haixia Han, Jiaqing Liang, Jie Shi, Qianyu He, and Yanghua Xiao. 2024.
\newblock \href {https://doi.org/10.1609/aaai.v38i16.29774} {Small language model can self-correct}.
\newblock In \emph{Proceedings of the AAAI Conference on Artificial Intelligence}, pages 18162--18170.

\bibitem[{He et~al.(2025)He, Polisetty, Fan, Huang, Wu, and Fung}]{he2025mmboundaryadvancingmllmknowledge}
Zhitao He, Sandeep Polisetty, Zhiyuan Fan, Yuchen Huang, Shujin Wu, and Yi~R. Fung. 2025.
\newblock \href {https://arxiv.org/abs/2505.23224} {Mmboundary: Advancing mllm knowledge boundary awareness through reasoning step confidence calibration}.
\newblock \emph{Preprint}, arXiv:2505.23224.

\bibitem[{Hu et~al.(2022)Hu, Shen, Wallis, Allen-Zhu, Li, Wang, Wang, and Chen}]{hu2022lora}
Edward~J Hu, Yelong Shen, Phillip Wallis, Zeyuan Allen-Zhu, Yuanzhi Li, Shean Wang, Lu~Wang, and Weizhu Chen. 2022.
\newblock \href {https://openreview.net/forum?id=nZeVKeeFYf9} {Lo{RA}: Low-rank adaptation of large language models}.
\newblock In \emph{International Conference on Learning Representations}.

\bibitem[{Huang et~al.(2023)Huang, Gu, Hou, Wu, Wang, Yu, and Han}]{huang2022large}
Jiaxin Huang, Shixiang~Shane Gu, Le~Hou, Yuexin Wu, Xuezhi Wang, Hongkun Yu, and Jiawei Han. 2023.
\newblock Large language models can self-improve.
\newblock In \emph{The 2023 Conference on Empirical Methods in Natural Language Processing}.

\bibitem[{Huang et~al.(2024{\natexlab{a}})Huang, Chen, Mishra, Zheng, Yu, Song, and Zhou}]{huang2023large}
Jie Huang, Xinyun Chen, Swaroop Mishra, Huaixiu~Steven Zheng, Adams~Wei Yu, Xinying Song, and Denny Zhou. 2024{\natexlab{a}}.
\newblock \href {https://openreview.net/forum?id=IkmD3fKBPQ} {Large language models cannot self-correct reasoning yet}.
\newblock In \emph{The Twelfth International Conference on Learning Representations}.

\bibitem[{Huang et~al.(2025{\natexlab{a}})Huang, Chen, Guo, Zeng, Zhao, Wu, Yuan, Zhao, Guo, Zhang, Yuan, Ju, Liu, Liu, Chang, and Zhang}]{huang-etal-2025-mmevalpro}
Jinsheng Huang, Liang Chen, Taian Guo, Fu~Zeng, Yusheng Zhao, Bohan Wu, Ye~Yuan, Haozhe Zhao, Zhihui Guo, Yichi Zhang, Jingyang Yuan, Wei Ju, Luchen Liu, Tianyu Liu, Baobao Chang, and Ming Zhang. 2025{\natexlab{a}}.
\newblock \href {https://aclanthology.org/2025.naacl-long.247/} {{MME}val{P}ro: Calibrating multimodal benchmarks towards trustworthy and efficient evaluation}.
\newblock In \emph{Proceedings of the 2025 Conference of the Nations of the Americas Chapter of the Association for Computational Linguistics: Human Language Technologies (Volume 1: Long Papers)}, pages 4805--4822, Albuquerque, New Mexico. Association for Computational Linguistics.

\bibitem[{Huang et~al.(2025{\natexlab{b}})Huang, He, Polisetty, Wang, and Fung}]{huang2025mactuningllmmulticompositionalproblem}
Junsheng Huang, Zhitao He, Sandeep Polisetty, Qingyun Wang, and May Fung. 2025{\natexlab{b}}.
\newblock \href {https://arxiv.org/abs/2504.21773} {Mac-tuning: Llm multi-compositional problem reasoning with enhanced knowledge boundary awareness}.
\newblock \emph{Preprint}, arXiv:2504.21773.

\bibitem[{Huang et~al.(2024{\natexlab{b}})Huang, Chan, Fung, Qiu, Zhou, Joty, Chang, and Ji}]{huang2024pixelsinsightssurveyautomatic}
Kung-Hsiang Huang, Hou~Pong Chan, Yi~R. Fung, Haoyi Qiu, Mingyang Zhou, Shafiq Joty, Shih-Fu Chang, and Heng Ji. 2024{\natexlab{b}}.
\newblock \href {https://arxiv.org/abs/2403.12027} {From pixels to insights: A survey on automatic chart understanding in the era of large foundation models}.
\newblock \emph{Preprint}, arXiv:2403.12027.

\bibitem[{Huang et~al.(2025{\natexlab{c}})Huang, Wang, Zhong, Su, Feng, Cao, and Fung}]{huang2025adactrladaptivecontrollablereasoning}
Shijue Huang, Hongru Wang, Wanjun Zhong, Zhaochen Su, Jiazhan Feng, Bowen Cao, and Yi~R. Fung. 2025{\natexlab{c}}.
\newblock \href {https://arxiv.org/abs/2505.18822} {Adactrl: Towards adaptive and controllable reasoning via difficulty-aware budgeting}.
\newblock \emph{Preprint}, arXiv:2505.18822.

\bibitem[{Kamoi et~al.(2024)Kamoi, Zhang, Zhang, Han, and Zhang}]{kamoi2024can}
Ryo Kamoi, Yusen Zhang, Nan Zhang, Jiawei Han, and Rui Zhang. 2024.
\newblock \href {https://doi.org/10.1162/tacl_a_00713} {When can {LLM}s actually correct their own mistakes? a critical survey of self-correction of {LLM}s}.
\newblock \emph{Transactions of the Association for Computational Linguistics}, 12:1417--1440.

\bibitem[{Kembhavi et~al.(2016)Kembhavi, Salvato, Kolve, Seo, Hajishirzi, and Farhadi}]{kembhavi2016diagram}
Aniruddha Kembhavi, Mike Salvato, Eric Kolve, Minjoon Seo, Hannaneh Hajishirzi, and Ali Farhadi. 2016.
\newblock \href {https://arxiv.org/abs/1603.07396} {A diagram is worth a dozen images}.
\newblock \emph{Preprint}, arXiv:1603.07396.

\bibitem[{Kumar et~al.(2024)Kumar, Zhuang, Agarwal, Su, Co-Reyes, Singh, Baumli, Iqbal, Bishop, Roelofs, Zhang, McKinney, Shrivastava, Paduraru, Tucker, Precup, Behbahani, and Faust}]{kumar2024traininglanguagemodelsselfcorrect}
Aviral Kumar, Vincent Zhuang, Rishabh Agarwal, Yi~Su, John~D Co-Reyes, Avi Singh, Kate Baumli, Shariq Iqbal, Colton Bishop, Rebecca Roelofs, Lei~M Zhang, Kay McKinney, Disha Shrivastava, Cosmin Paduraru, George Tucker, Doina Precup, Feryal Behbahani, and Aleksandra Faust. 2024.
\newblock \href {https://arxiv.org/abs/2409.12917} {Training language models to self-correct via reinforcement learning}.
\newblock \emph{Preprint}, arXiv:2409.12917.

\bibitem[{Li et~al.(2023{\natexlab{a}})Li, Wang, Wang, Ge, Ge, and Shan}]{li2023seed}
Bohao Li, Rui Wang, Guangzhi Wang, Yuying Ge, Yixiao Ge, and Ying Shan. 2023{\natexlab{a}}.
\newblock \href {https://arxiv.org/abs/2307.16125} {Seed-bench: Benchmarking multimodal llms with generative comprehension}.
\newblock \emph{Preprint}, arXiv:2307.16125.

\bibitem[{Li et~al.(2025)Li, Fung, Wang, Han, Li, Wang, and Ji}]{li2025mentalarenaselfplaytraininglanguage}
Cheng Li, May Fung, Qingyun Wang, Chi Han, Manling Li, Jindong Wang, and Heng Ji. 2025.
\newblock \href {https://arxiv.org/abs/2410.06845} {Mentalarena: Self-play training of language models for diagnosis and treatment of mental health disorders}.
\newblock \emph{Preprint}, arXiv:2410.06845.

\bibitem[{Li et~al.(2023{\natexlab{b}})Li, Xie, Li, Chen, Wang, Chen, Yang, Wang, and Kong}]{li2023silkiepreferencedistillationlarge}
Lei Li, Zhihui Xie, Mukai Li, Shunian Chen, Peiyi Wang, Liang Chen, Yazheng Yang, Benyou Wang, and Lingpeng Kong. 2023{\natexlab{b}}.
\newblock \href {https://arxiv.org/abs/2312.10665} {Silkie: Preference distillation for large visual language models}.
\newblock \emph{Preprint}, arXiv:2312.10665.

\bibitem[{Li et~al.(2024)Li, Chen, Chen, Zhang, Su, Xing, and Zhang}]{li2024confidence}
Loka Li, Zhenhao Chen, Guangyi Chen, Yixuan Zhang, Yusheng Su, Eric Xing, and Kun Zhang. 2024.
\newblock \href {https://arxiv.org/abs/2402.12563} {Confidence matters: Revisiting intrinsic self-correction capabilities of large language models}.
\newblock \emph{Preprint}, arXiv:2402.12563.

\bibitem[{Li et~al.(2023{\natexlab{c}})Li, Klavins, Xu, Zafri, and Stern}]{li2023understandingdriverpedestrianinteractionspredict}
Tianyi Li, Joshua Klavins, Te~Xu, Niaz~Mahmud Zafri, and Raphael Stern. 2023{\natexlab{c}}.
\newblock \href {https://arxiv.org/abs/2312.15113} {Understanding driver-pedestrian interactions to predict driver yielding: naturalistic open-source dataset collected in minnesota}.
\newblock \emph{Preprint}, arXiv:2312.15113.

\bibitem[{Liu et~al.(2024{\natexlab{a}})Liu, Nassereldine, Yang, Xu, Hu, Li, Kumar, Lee, Qin, Shi, and Xiong}]{liu2024largelanguagemodelsintrinsic}
Dancheng Liu, Amir Nassereldine, Ziming Yang, Chenhui Xu, Yuting Hu, Jiajie Li, Utkarsh Kumar, Changjae Lee, Ruiyang Qin, Yiyu Shi, and Jinjun Xiong. 2024{\natexlab{a}}.
\newblock \href {https://arxiv.org/abs/2406.15673} {Large language models have intrinsic self-correction ability}.
\newblock \emph{Preprint}, arXiv:2406.15673.

\bibitem[{Liu et~al.(2024{\natexlab{b}})Liu, Li, Li, and Lee}]{liu2024improved}
Haotian Liu, Chunyuan Li, Yuheng Li, and Yong~Jae Lee. 2024{\natexlab{b}}.
\newblock Improved baselines with visual instruction tuning.
\newblock In \emph{Proceedings of the IEEE/CVF Conference on Computer Vision and Pattern Recognition (CVPR)}, pages 26296--26306.

\bibitem[{Liu et~al.(2024{\natexlab{c}})Liu, Duan, Zhang, Li, Zhang, Zhao, Yuan, Wang, He, Liu, Chen, and Lin}]{10.1007/978-3-031-72658-3_13}
Yuan Liu, Haodong Duan, Yuanhan Zhang, Bo~Li, Songyang Zhang, Wangbo Zhao, Yike Yuan, Jiaqi Wang, Conghui He, Ziwei Liu, Kai Chen, and Dahua Lin. 2024{\natexlab{c}}.
\newblock \href {https://doi.org/10.1007/978-3-031-72658-3_13} {Mmbench: Is your multi-modal model an all-around player?}
\newblock In \emph{Computer Vision – ECCV 2024: 18th European Conference, Milan, Italy, September 29–October 4, 2024, Proceedings, Part VI}, page 216–233, Berlin, Heidelberg. Springer-Verlag.

\bibitem[{Lu et~al.(2022)Lu, Mishra, Xia, Qiu, Chang, Zhu, Tafjord, Clark, and Kalyan}]{lu2022learn}
Pan Lu, Swaroop Mishra, Tanglin Xia, Liang Qiu, Kai-Wei Chang, Song-Chun Zhu, Oyvind Tafjord, Peter Clark, and Ashwin Kalyan. 2022.
\newblock \href {https://proceedings.neurips.cc/paper_files/paper/2022/file/11332b6b6cf4485b84afadb1352d3a9a-Paper-Conference.pdf} {Learn to explain: Multimodal reasoning via thought chains for science question answering}.
\newblock In \emph{Advances in Neural Information Processing Systems}, volume~35, pages 2507--2521. Curran Associates, Inc.

\bibitem[{Madaan et~al.(2023)Madaan, Tandon, Gupta, Hallinan, Gao, Wiegreffe, Alon, Dziri, Prabhumoye, Yang, Gupta, Majumder, Hermann, Welleck, Yazdanbakhsh, and Clark}]{madaan2024self}
Aman Madaan, Niket Tandon, Prakhar Gupta, Skyler Hallinan, Luyu Gao, Sarah Wiegreffe, Uri Alon, Nouha Dziri, Shrimai Prabhumoye, Yiming Yang, Shashank Gupta, Bodhisattwa~Prasad Majumder, Katherine Hermann, Sean Welleck, Amir Yazdanbakhsh, and Peter Clark. 2023.
\newblock \href {https://proceedings.neurips.cc/paper_files/paper/2023/file/91edff07232fb1b55a505a9e9f6c0ff3-Paper-Conference.pdf} {Self-refine: Iterative refinement with self-feedback}.
\newblock In \emph{Advances in Neural Information Processing Systems}, volume~36, pages 46534--46594. Curran Associates, Inc.

\bibitem[{OpenAI(2024{\natexlab{a}})}]{gpt-4o}
OpenAI. 2024{\natexlab{a}}.
\newblock Hello gpt-4o.
\newblock \url{https://openai.com/index/hello-gpt-4o/}.
\newblock Accessed: 2025-05-29.

\bibitem[{OpenAI(2024{\natexlab{b}})}]{gpt-o1}
OpenAI. 2024{\natexlab{b}}.
\newblock O1 system card.
\newblock \url{https://cdn.openai. com/o1-system-card-20240917.pdf}.
\newblock Accessed: 2025-05-29.

\bibitem[{Peng et~al.(2024)Peng, Zhang, Yang, Zhang, and Qiao}]{peng2024data}
Wenshuo Peng, Kaipeng Zhang, Yue Yang, Hao Zhang, and Yu~Qiao. 2024.
\newblock Data adaptive traceback for vision-language foundation models in image classification.
\newblock In \emph{Proceedings of the AAAI Conference on Artificial Intelligence}, volume~38, pages 4506--4514.

\bibitem[{Phute et~al.(2024)Phute, Helbling, Hull, Peng, Szyller, Cornelius, and Chau}]{helbling2023llm}
Mansi Phute, Alec Helbling, Matthew~Daniel Hull, ShengYun Peng, Sebastian Szyller, Cory Cornelius, and Duen~Horng Chau. 2024.
\newblock \href {https://openreview.net/forum?id=YoqgcIA19o} {{LLM} self defense: By self examination, {LLM}s know they are being tricked}.
\newblock In \emph{The Second Tiny Papers Track at ICLR 2024}.

\bibitem[{Rafailov et~al.(2023)Rafailov, Sharma, Mitchell, Manning, Ermon, and Finn}]{rafailov2024direct}
Rafael Rafailov, Archit Sharma, Eric Mitchell, Christopher~D Manning, Stefano Ermon, and Chelsea Finn. 2023.
\newblock \href {https://proceedings.neurips.cc/paper_files/paper/2023/file/a85b405ed65c6477a4fe8302b5e06ce7-Paper-Conference.pdf} {Direct preference optimization: Your language model is secretly a reward model}.
\newblock In \emph{Advances in Neural Information Processing Systems}, volume~36, pages 53728--53741. Curran Associates, Inc.

\bibitem[{Shinn et~al.(2023)Shinn, Cassano, Gopinath, Narasimhan, and Yao}]{shinn2024reflexion}
Noah Shinn, Federico Cassano, Ashwin Gopinath, Karthik Narasimhan, and Shunyu Yao. 2023.
\newblock \href {https://proceedings.neurips.cc/paper_files/paper/2023/file/1b44b878bb782e6954cd888628510e90-Paper-Conference.pdf} {Reflexion: language agents with verbal reinforcement learning}.
\newblock In \emph{Advances in Neural Information Processing Systems}, volume~36, pages 8634--8652. Curran Associates, Inc.

\bibitem[{Sun et~al.(2025)Sun, Yang, Reddy, Fung, Chan, Small, Zhai, and Ji}]{sun2025personadbefficientlargelanguage}
Chenkai Sun, Ke~Yang, Revanth~Gangi Reddy, Yi~R. Fung, Hou~Pong Chan, Kevin Small, ChengXiang Zhai, and Heng Ji. 2025.
\newblock \href {https://arxiv.org/abs/2402.11060} {Persona-db: Efficient large language model personalization for response prediction with collaborative data refinement}.
\newblock \emph{Preprint}, arXiv:2402.11060.

\bibitem[{Tong et~al.(2024)Tong, Li, Wang, Wang, Teng, and Shang}]{tong2024can}
Yongqi Tong, Dawei Li, Sizhe Wang, Yujia Wang, Fei Teng, and Jingbo Shang. 2024.
\newblock \href {https://doi.org/10.18653/v1/2024.acl-long.169} {Can {LLM}s learn from previous mistakes? investigating {LLM}s' errors to boost for reasoning}.
\newblock In \emph{Proceedings of the 62nd Annual Meeting of the Association for Computational Linguistics (Volume 1: Long Papers)}, pages 3065--3080, Bangkok, Thailand. Association for Computational Linguistics.

\bibitem[{Wang et~al.(2025{\natexlab{a}})Wang, Chen, Wang, Zhou, Zhou, Yao, Zhou, Goldstein, Bhatia, Kass-Hout, Huang, and Xiao}]{wang-etal-2025-enhancing}
Xiyao Wang, Jiuhai Chen, Zhaoyang Wang, Yuhang Zhou, Yiyang Zhou, Huaxiu Yao, Tianyi Zhou, Tom Goldstein, Parminder Bhatia, Taha Kass-Hout, Furong Huang, and Cao Xiao. 2025{\natexlab{a}}.
\newblock \href {https://aclanthology.org/2025.findings-naacl.15/} {Enhancing visual-language modality alignment in large vision language models via self-improvement}.
\newblock In \emph{Findings of the Association for Computational Linguistics: NAACL 2025}, pages 268--282, Albuquerque, New Mexico. Association for Computational Linguistics.

\bibitem[{Wang et~al.(2023)Wang, Wei, Schuurmans, Le, Chi, Narang, Chowdhery, and Zhou}]{wang2022self}
Xuezhi Wang, Jason Wei, Dale Schuurmans, Quoc~V Le, Ed~H. Chi, Sharan Narang, Aakanksha Chowdhery, and Denny Zhou. 2023.
\newblock \href {https://openreview.net/forum?id=1PL1NIMMrw} {Self-consistency improves chain of thought reasoning in language models}.
\newblock In \emph{The Eleventh ICLR}.

\bibitem[{Wang et~al.(2025{\natexlab{b}})Wang, Fan, Wang, Fung, and Ji}]{wang2025calmunleashingcrosslingualselfaligning}
Yumeng Wang, Zhiyuan Fan, Qingyun Wang, May Fung, and Heng Ji. 2025{\natexlab{b}}.
\newblock \href {https://arxiv.org/abs/2501.18457} {Calm: Unleashing the cross-lingual self-aligning ability of language model question answering}.
\newblock \emph{Preprint}, arXiv:2501.18457.

\bibitem[{Wu et~al.(2024{\natexlab{a}})Wu, Fung, Qian, Kim, Hakkani-Tur, and Ji}]{wu2024aligningllmsindividualpreferences}
Shujin Wu, May Fung, Cheng Qian, Jeonghwan Kim, Dilek Hakkani-Tur, and Heng Ji. 2024{\natexlab{a}}.
\newblock \href {https://arxiv.org/abs/2410.03642} {Aligning llms with individual preferences via interaction}.
\newblock \emph{Preprint}, arXiv:2410.03642.

\bibitem[{Wu et~al.(2024{\natexlab{b}})Wu, Fung, Li, Wan, Chang, and Ji}]{wu-etal-2024-macaroon}
Shujin Wu, Yi~Fung, Sha Li, Yixin Wan, Kai-Wei Chang, and Heng Ji. 2024{\natexlab{b}}.
\newblock \href {https://doi.org/10.18653/v1/2024.findings-emnlp.454} {{MACAROON}: Training vision-language models to be your engaged partners}.
\newblock In \emph{Findings of the Association for Computational Linguistics: EMNLP 2024}, pages 7715--7731, Miami, Florida, USA. Association for Computational Linguistics.

\bibitem[{xAI(2024)}]{RealWorldQA}
xAI. 2024.
\newblock Realworldqa.
\newblock \url{https://x.ai/blog/grok-1.5v}.
\newblock Accessed: 2025-05-29.

\bibitem[{Xu et~al.(2024)Xu, Zhu, Zhao, Pan, Li, and Wang}]{xu-etal-2024-pride}
Wenda Xu, Guanglei Zhu, Xuandong Zhao, Liangming Pan, Lei Li, and William Wang. 2024.
\newblock \href {https://doi.org/10.18653/v1/2024.acl-long.826} {Pride and prejudice: {LLM} amplifies self-bias in self-refinement}.
\newblock In \emph{Proceedings of the 62nd Annual Meeting of the Association for Computational Linguistics (Volume 1: Long Papers)}, pages 15474--15492, Bangkok, Thailand. Association for Computational Linguistics.

\bibitem[{Yang et~al.(2024)Yang, Chen, Rao, Guo, Zhang, Yang, and Zhang}]{yang2024tackling}
Diji Yang, Kezhen Chen, Jinmeng Rao, Xiaoyuan Guo, Yawen Zhang, Jie Yang, and Yi~Zhang. 2024.
\newblock Tackling vision language tasks through learning inner monologues.
\newblock In \emph{Proceedings of the AAAI Conference on Artificial Intelligence}, volume~38, pages 19350--19358.

\bibitem[{Yao et~al.(2024)Yao, Yu, Zhang, Wang, Cui, Zhu, Cai, Li, Zhao, He, Chen, Zhou, Zou, Zhang, Hu, Zheng, Zhou, Cai, Han, Zeng, Li, Liu, and Sun}]{yao2024minicpm}
Yuan Yao, Tianyu Yu, Ao~Zhang, Chongyi Wang, Junbo Cui, Hongji Zhu, Tianchi Cai, Haoyu Li, Weilin Zhao, Zhihui He, Qianyu Chen, Huarong Zhou, Zhensheng Zou, Haoye Zhang, Shengding Hu, Zhi Zheng, Jie Zhou, Jie Cai, Xu~Han, and 4 others. 2024.
\newblock \href {https://arxiv.org/abs/2408.01800} {Minicpm-v: A gpt-4v level mllm on your phone}.
\newblock \emph{Preprint}, arXiv:2408.01800.

\bibitem[{Ying et~al.(2024)Ying, Meng, Wang, Li, Lin, Yang, Zhang, Zhang, Lin, Liu, jiayi lei, Lu, Chen, Xu, Zhang, Zhang, Gao, Wang, Qiao, Luo, Zhang, and Shao}]{ying2024mmtbench}
Kaining Ying, Fanqing Meng, Jin Wang, Zhiqian Li, Han Lin, Yue Yang, Hao Zhang, Wenbo Zhang, Yuqi Lin, Shuo Liu, jiayi lei, Quanfeng Lu, Runjian Chen, Peng Xu, Renrui Zhang, Haozhe Zhang, Peng Gao, Yali Wang, Yu~Qiao, and 3 others. 2024.
\newblock \href {https://openreview.net/forum?id=R4Ng8zYaiz} {{MMT}-bench: A comprehensive multimodal benchmark for evaluating large vision-language models towards multitask {AGI}}.
\newblock In \emph{Forty-first International Conference on Machine Learning}.

\bibitem[{Yue et~al.(2024)Yue, Ni, Zhang, Zheng, Liu, Zhang, Stevens, Jiang, Ren, Sun, Wei, Yu, Yuan, Sun, Yin, Zheng, Yang, Liu, Huang, Sun, Su, and Chen}]{yue2023mmmu}
Xiang Yue, Yuansheng Ni, Kai Zhang, Tianyu Zheng, Ruoqi Liu, Ge~Zhang, Samuel Stevens, Dongfu Jiang, Weiming Ren, Yuxuan Sun, Cong Wei, Botao Yu, Ruibin Yuan, Renliang Sun, Ming Yin, Boyuan Zheng, Zhenzhu Yang, Yibo Liu, Wenhao Huang, and 3 others. 2024.
\newblock Mmmu: A massive multi-discipline multimodal understanding and reasoning benchmark for expert agi.
\newblock In \emph{Proceedings of the IEEE/CVF Conference on Computer Vision and Pattern Recognition (CVPR)}, pages 9556--9567.

\bibitem[{Zhang et~al.(2025)Zhang, Yao, Pi, Liang, and Fung}]{zhang2025vlm2benchcloserlookvlms}
Jianshu Zhang, Dongyu Yao, Renjie Pi, Paul~Pu Liang, and Yi~R. Fung. 2025.
\newblock \href {https://arxiv.org/abs/2502.12084} {Vlm2-bench: A closer look at how well vlms implicitly link explicit matching visual cues}.
\newblock \emph{Preprint}, arXiv:2502.12084.

\bibitem[{Zhang et~al.(2024)Zhang, Peng, Tian, Zhou, Jin, Song, Mi, and Meng}]{zhang2024self}
Xiaoying Zhang, Baolin Peng, Ye~Tian, Jingyan Zhou, Lifeng Jin, Linfeng Song, Haitao Mi, and Helen Meng. 2024.
\newblock \href {https://aclanthology.org/2024.acl-long.107} {Self-alignment for factuality: Mitigating hallucinations in {LLM}s via self-evaluation}.
\newblock In \emph{Proceedings of the 62nd Annual Meeting of the Association for Computational Linguistics (Volume 1: Long Papers)}, pages 1946--1965, Bangkok, Thailand. Association for Computational Linguistics.

\bibitem[{Zhou et~al.(2024{\natexlab{a}})Zhou, Cui, Rafailov, Finn, and Yao}]{zhou2024aligning}
Yiyang Zhou, Chenhang Cui, Rafael Rafailov, Chelsea Finn, and Huaxiu Yao. 2024{\natexlab{a}}.
\newblock \href {https://openreview.net/forum?id=GRGvC0rpA8} {Aligning modalities in vision large language models via preference fine-tuning}.
\newblock In \emph{ICLR 2024 Workshop on Reliable and Responsible Foundation Models}.

\bibitem[{Zhou et~al.(2024{\natexlab{b}})Zhou, Fan, Cheng, Yang, Chen, Cui, Wang, Li, Zhang, and Yao}]{zhou2024calibrated}
Yiyang Zhou, Zhiyuan Fan, Dongjie Cheng, Sihan Yang, Zhaorun Chen, Chenhang Cui, Xiyao Wang, Yun Li, Linjun Zhang, and Huaxiu Yao. 2024{\natexlab{b}}.
\newblock \href {https://openreview.net/forum?id=B7sj10tXbL} {Calibrated self-rewarding vision language models}.
\newblock In \emph{ICML 2024 Workshop on Foundation Models in the Wild}.

\bibitem[{Zhu et~al.(2024)Zhu, Chen, Shen, Li, and Elhoseiny}]{zhu2024minigpt}
Deyao Zhu, Jun Chen, Xiaoqian Shen, Xiang Li, and Mohamed Elhoseiny. 2024.
\newblock \href {https://openreview.net/forum?id=1tZbq88f27} {Mini{GPT}-4: Enhancing vision-language understanding with advanced large language models}.
\newblock In \emph{The Twelfth International Conference on Learning Representations}.

\end{thebibliography}

\appendix

\section{Data Examples of \datasetname{}}\label{Data Examples}

Our work introduces \datasetname{}, a novel dataset derived from the intrinsic self-correction processes of Vision-Language Models (VLMs) during inference. For each sample, the dataset captures both the VLM's initial and refined responses. Specifically, \datasetname{} comprises Type 2 (incorrect$\Rightarrow$correct) and Type 3 (correct$\Rightarrow$incorrect) self-correction instances, where the correct responses are designated as preferred and the incorrect ones as disfavored. In this section, we provide illustrative examples of the dataset content (Table~\ref{tab:LLaVA-V1.5-7B-type2} to \ref{tab:MiniCPM-Llama3-V-2.5-type3}).

\section{Examples Supporting Further Studies}\label{Examples Supporting Further Studies}

In our further studies, we argue that VLMs often exhibit limited capabilities for genuinely accurate refinement, and that even seemingly successful self-corrections can be unreliable due to flawed underlying reasoning. To substantiate this claim, we present two illustrative examples in Table~\ref{tab:further}.

In the first example, the model's refined response relies on the misleading assumption that a ``Do not board'' sign directly indicates a lack of crowding on the platform. However, platform crowding should primarily be assessed based on crowd density. Although the model ultimately selects the correct answer (B. no), this decision stems from an over-reliance on a single, potentially misleading cue rather than a comprehensive understanding of the entire visual scenario.

In the second example, the model confuses particle size with concentration during the refinement process. The model infers concentration levels from the size of the purple particles, entirely neglecting the actual number of particles present. While the model successfully corrects the error in its initial response, the self-correction process should have involved a proper reassessment of the quantity of purple particles in the solution, not their individual size.

\begin{table}[tbh]
    \centering
    \small
    \begin{tabular}{p{8cm}}
        \toprule
        \textbf{Q:} What is the direction of this pull? A. toward her hand B. away from her hand\\
        \textbf{I:} \includegraphics[width=0.3\linewidth]{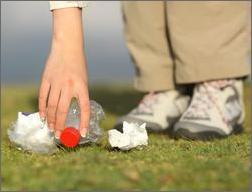} \\
        \textbf{IR (disfavored):} \textbf{B. away from her hand.} The person is pulling a bottle away from their hand, which indicates that they are moving the bottle in the opposite direction of their hand. This action suggests that they are trying to remove the bottle from the ground or from their hand, rather than placing it there.\\
        \textbf{RR (preferred):} \textbf{A. toward her hand.} Upon reviewing the image, it is clear that the person is pulling a bottle toward their hand, rather than away from it. This action suggests that they are trying to pick up the bottle or place it on the ground. The person's hand is positioned above the bottle, indicating that they are in the process of lifting it.\\
        \midrule
        \textbf{Q:} Detect and give a judgment of the anomalous object within the given image. A. normal B. anomaly\\
        \textbf{I:} \includegraphics[width=0.4\linewidth]{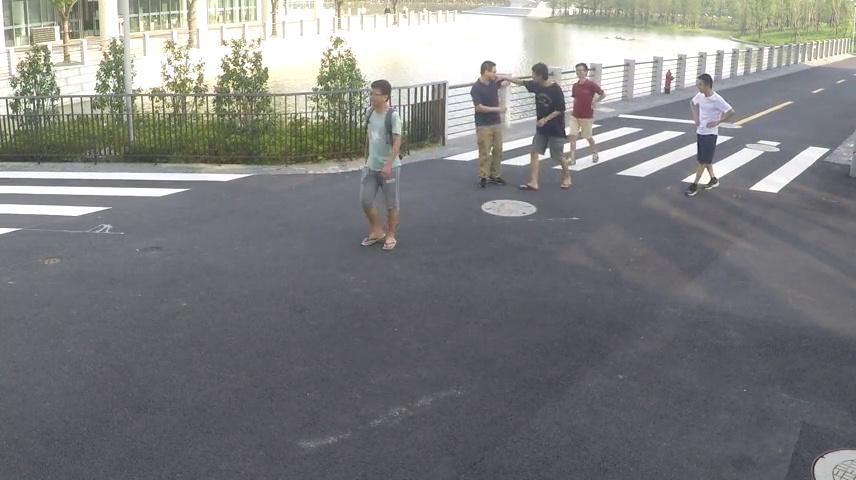} \\
        \textbf{IR (disfavored):} \textbf{B. anomaly.} The image shows a group of people walking on a sidewalk near a body of water. Among them, there is a person wearing a backpack, which is an unusual object in this context. The backpack is not a typical item one would expect to see while walking on a sidewalk, as it is usually associated with carrying personal belongings or items for outdoor activities. The presence of the backpack in this scene might indicate that the person is either on a trip or carrying something specific, but it is not a common sight in this setting.\\
        \textbf{RR (preferred):} \textbf{A. normal.} After reviewing the image, I realized that the backpack is not an anomalous object in this context. It is a common item that people carry while walking on a sidewalk, especially if they are on a trip or carrying personal belongings. The presence of the backpack does not indicate that the person is carrying something specific or that they are on a trip. Therefore, the backpack is not an anomalous object in this scene.\\
        \bottomrule
    \end{tabular}
    \caption{Type 2 samples of LLaVA-V1.5-7B.}
    \label{tab:LLaVA-V1.5-7B-type2}
\end{table}

\begin{table}[tbh]
    \centering
    \small
    \begin{tabular}{p{8cm}}
        \toprule
        \textbf{Q:} How would you describe the appearance of the man playing the drums? A. Young and long-haired B. Middle-aged with glasses C. Old and bald D. Young with a beard\\
        \textbf{I:} \includegraphics[width=0.25\linewidth]{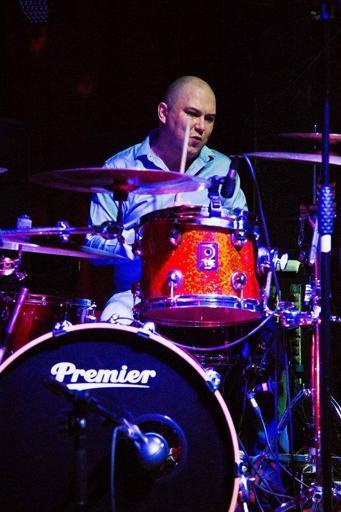} \\
        \textbf{IR (preferred):} \textbf{C. Old and bald.} The man playing the drums is described as an older man with a bald head. This suggests that he is not a young man, and his appearance is not typical of a young musician. The fact that he is playing drums indicates that he has some level of skill and experience in the field.\\
        \textbf{RR (disfavored):} \textbf{A. Young and long-haired.} Upon reviewing the image, it is clear that the man playing the drums is not an older man with a bald head. Instead, he is described as a young man with long hair. This suggests that he is not a middle-aged man, and his appearance is more typical of a young musician. The fact that he is playing drums indicates that he has some level of skill and experience in the field\\
        \midrule
        \textbf{Q:} What landmark is this and where is it? A. Mecca in Saudi Arabia B. Big Ben in London C. The Burj al Arab Hotel in Dubai D. Tower of Pisa, Italy \\
        \textbf{I:} \includegraphics[width=0.25\linewidth]{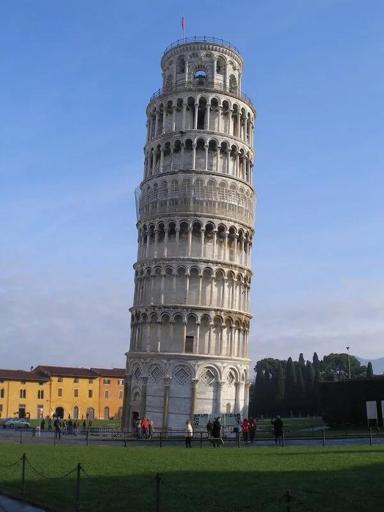} \\
        \textbf{IR (preferred):} \textbf{D. Tower of Pisa, Italy.} The image features a tall, white tower with a clock on its side, which is the iconic Tower of Pisa. The tower is located in Italy, and it is a famous landmark known for its leaning structure. The presence of a large group of people around the tower suggests that it is a popular tourist attraction. The image also shows a clock on the side of the tower, which is a distinctive feature of the structure.\\
        \textbf{RR (disfavored):} \textbf{A. Mecca in Saudi Arabia.} I apologize for the confusion in my previous response. After reviewing the image, I realized that the Tower of Pisa is not the correct answer. The image features a large clock tower, which is a prominent landmark in Mecca, Saudi Arabia. The tower is known as the Abraj Al-Bait Clock Tower and is a significant attraction for pilgrims visiting the city. The image shows a large group of people around the tower, indicating its popularity as a tourist destination. The clock tower's distinctive design and its location in Mecca make it the most appropriate landmark in the image.\\
        \bottomrule
    \end{tabular}
    \caption{Type 3 samples of LLaVA-V1.5-7B.}
    \label{tab:LLaVA-V1.5-7B-type3}
\end{table}

\begin{table}[t]
    \centering
    \small
    \begin{tabular}{p{8cm}}
        \toprule
        \textbf{Q:} How many arrows are there in total in the picture? A. 1 B. 2 C. 3 D. 4\\
        \textbf{I:} \includegraphics[width=0.5\linewidth]{figure/arrow.jpg} \\
        \textbf{IR (disfavored):} There are a total of \textbf{3 arrows} in the picture. The first arrow is pointing upwards, the second arrow is pointing downwards, and the third arrow is pointing to the right. \\
        \textbf{RR (preferred):} Upon reviewing my previous answer, I realized that I missed the fact that there are two arrows pointing in the same direction, which is upwards. Therefore, there are a total of \textbf{4 arrows} in the picture. The first arrow is pointing upwards, the second arrow is pointing upwards as well, the third arrow is pointing to the right, and the fourth arrow is pointing downwards.\\
        \midrule
        \textbf{Q:} How many species are there in total? A. 5 B. 6 C. 7 D. 8 E. 9\\
        \textbf{I:} \includegraphics[width=0.7\linewidth]{figure/species.jpg} \\
        \textbf{IR (disfavored):} There are a total of \textbf{7 species} in the diagram. The diagram shows the food chain of a water ecosystem, with each species being connected to the next one through arrows. The species include algae, water flea, tadpole, great diving beetle, and tapioca. The arrows indicate the flow of energy and nutrients from one species to another, as each species consumes the one before it.\\
        \textbf{RR (preferred):} Upon reviewing my previous answer, I realized that I missed one species in the food chain. The correct answer is \textbf{8 species.} The diagram shows the food chain of a water ecosystem, with each species being connected to the next one through arrows. The species include algae, water flea, tadpole, great diving beetle, tapioca, and two other species that are not explicitly identified. The arrows indicate the flow of energy and nutrients from one species to another, as each species consumes the one before it.\\
        \bottomrule
    \end{tabular}
    \caption{Type 2 samples of LLaVA-V1.5-13B.}
    \label{tab:LLaVA-V1.5-13B-type2}
\end{table}

\begin{table}[t]
    \centering
    \small
    \begin{tabular}{p{8cm}}
        \toprule
        \textbf{Q:} Where do Teal and Dark Khaki intersect on the x-axis? A. 40 B. 44 C. 60 D. 80\\
        \textbf{I:} \includegraphics[width=0.5\linewidth]{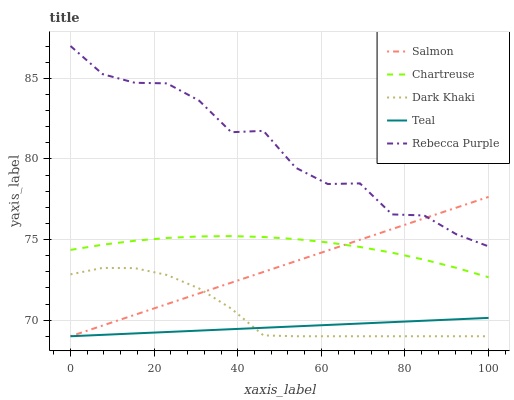} \\
        \textbf{IR (preferred):} To find the intersection of Teal and Dark Khaki on the x-axis, we need to look at the x-axis values for both lines. Teal has a value of 40, while Dark Khaki has a value of 60. The intersection occurs at the point where these two values meet, which is at 44. Therefore, the answer is \textbf{B. 44.}\\
        \textbf{RR (disfavored):} Upon reviewing my previous answer, I realized that I made an error in my interpretation of the x-axis values for Teal and Dark Khaki. Teal has a value of 40, while Dark Khaki has a value of 80. The intersection occurs at the point where these two values meet, which is at 60. Therefore, the correct answer is \textbf{C. 60.}\\
        \midrule
        \textbf{Q:} What color is the Howler Monkey? A. Blue B. Green C. Brown D. Purple E. Black F. Dark brown G. White\\
        \textbf{I:} \includegraphics[width=0.4\linewidth]{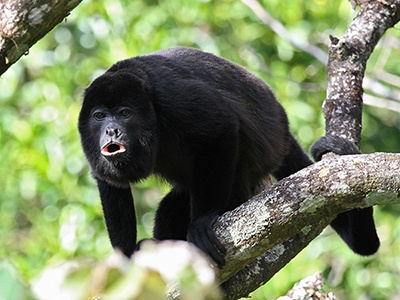} \\
        \textbf{IR (preferred):} The Howler Monkey in the image is black. The monkey is sitting on a tree branch, and its color is clearly visible. The monkey's black fur stands out against the tree branch, making it easy to identify the color of the monkey. Therefore, the correct answer is \textbf{E. Black.}\\
        \textbf{RR (disfavored):} Upon reviewing my previous answer, I noticed that I did not consider the possibility of the monkey's fur appearing dark brown in certain lighting conditions. Therefore, I would like to revise my answer to include the possibility of the monkey's fur appearing dark brown or brown. Therefore, the correct answer is \textbf{F. Dark brown.}\\
        \bottomrule
    \end{tabular}
    \caption{Type 3 samples of LLaVA-V1.5-13B.}
    \label{tab:LLaVA-V1.5-13B-type3}
\end{table}

\begin{table}[t]
    \centering
    \small
    \begin{tabular}{p{8cm}}
        \toprule
        \textbf{Q:} Select the amphibian below. 0: brown tree frog 1: woodpecker\\
        \textbf{I:} \includegraphics[width=0.35\linewidth]{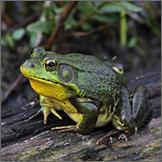} \\
        \textbf{IR (disfavored):} The image features a frog, which is an amphibian. The brown tree frog (option 0) is also an amphibian, but it is not depicted in the image; therefore, by elimination and based on the visual characteristics of the animal in the picture, the correct answer is \textbf{1 (woodpecker)}, as it does not match the description of the amphibian shown. \\
        \textbf{RR (preferred):} Upon reviewing the context, it's clear that the image showcases a frog, which is an amphibian. The correct answer should be the amphibian present in the image. Therefore, the improved answer is \textbf{0 (brown tree frog)}, as it matches the description of the amphibian depicted in the image.\\
        \midrule
        \textbf{Q:} Which solution has a higher concentration of green particles? A. neither; their concentrations are the same B. Solution B C. Solution A\\
        \textbf{I:} \includegraphics[width=0.45\linewidth]{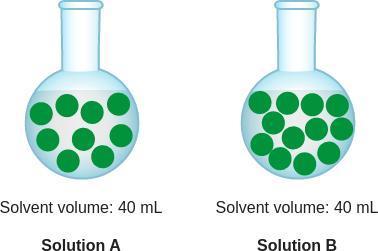} \\
        \textbf{IR (disfavored):} The concentration of particles in a solution is determined by the volume of the solution and the number of particles present. In this case, both Solution A and Solution B have a solvent volume of 40 ml. Since there are more green particles in Solution A than in Solution B, we can conclude that Solution A has a higher concentration of green particles. Final answer: \textbf{C. Solution A}\\
        \textbf{RR (preferred)}: After reviewing the image, it is clear that there are more green particles in Solution B than in Solution A. Therefore, the concentration of green particles is higher in Solution B. Final answer: \textbf{B. Solution B} \\
        \bottomrule
    \end{tabular}
    \caption{Type 2 samples of MiniCPM-Llama3-V-2.5.}
    \label{tab:MiniCPM-Llama3-V-2.5-type2}
\end{table}

\begin{table}[t]
    \centering
    \small
    \begin{tabular}{p{8cm}}
        \toprule
        \textbf{Q:} Which term matches the picture? A. endotherms B. ectotherms\\
        \textbf{I:}\includegraphics[width=0.4\linewidth]{figure/horse.jpg} \\
        \textbf{IR (preferred):} The picture you provided shows a horse, which is a mammal. Mammals are endotherms, meaning they can regulate their body temperature internally. The term that matches the picture is \textbf{A. endotherms}.\\
        \textbf{RR (disfavored):} I have reviewed my previous answer and considered all relevant aspects of the image. Based on my review, I believe that the term ectotherms is more suitable to describe the horse in the picture because horses are cold-blooded animals that regulate their body temperature by external heat sources, rather than internally like endotherms (warm-blooded animals). Therefore, my final answer is \textbf{B. ectotherms}.\\
        \midrule
        \textbf{Q:} Which term matches the picture? A. Hawaiian eruption B. Strombolian eruption\\
        \textbf{I:}\includegraphics[width=0.4\linewidth]{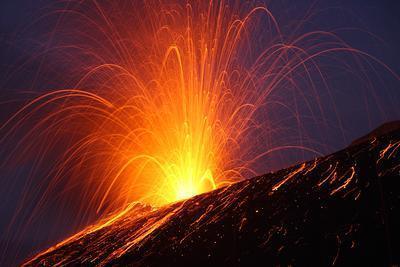} \\
        \textbf{IR (preferred):} A Strombolian eruption is characterized by moderate bursts of expanding gases that eject incandescent lava fragments, as shown in the image with fiery lava being thrown into the air. Hawaiian eruptions, on the other hand, are generally less explosive and produce more fluid lava flows. The term that matches the picture is \textbf{B. Strombolian eruption}.\\
        \textbf{RR (disfavored):} I apologize for the mistake in my previous answer. Upon reviewing the image, it is clear that the term that matches the picture is \textbf{A. Hawaiian eruption}. This is because the image shows a volcanic eruption with lava flowing and a bright light at the top of the volcano, which are characteristics of a Hawaiian eruption. Strombolian eruptions typically involve explosive ejections of solid material and do not necessarily exhibit such visible lava flows or bright lights as seen in the image. Therefore, I have corrected my initial response to accurately reflect the information provided in the image.\\
        \bottomrule
    \end{tabular}
    \caption{Type 3 samples of MiniCPM-Llama3-V-2.5.}
    \label{tab:MiniCPM-Llama3-V-2.5-type3}
\end{table}

\begin{table}[t]
    \centering
    \small
    \begin{tabular}{p{8cm}}
        \toprule
        \textbf{Q:} is this place crowded? A. yes B. no\\
        \textbf{I:} \includegraphics[width=0.4\linewidth]{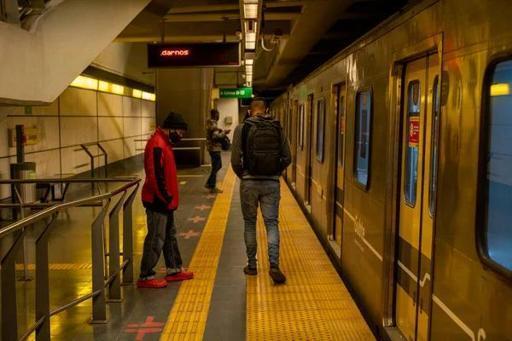} \\
        \textbf{IR:} The place is crowded. People are waiting for the train, and they are spread out along the platform. The final answer is \textbf{A}.\\
        \textbf{SC:} Review your previous answer and ensure that all relevant aspects of the image have been considered. Are there any elements or details that you missed? Based on your review, improve your answer. \\
        \textbf{RR:} I reviewed my previous answer and have identified a relevant detail that was missed. The sign Do not board on the platform indicates that the train may be not crowded, as it is likely the reason why people are standing near it. Based on this new information, I will revise my final answer to \textbf{B}.\\
        \midrule
        \textbf{Q:} Which solution has a higher concentration of purple particles? A. Solution A B. Solution B C. neither; their concentrations are the same\\
        \textbf{I:} \includegraphics[width=0.4\linewidth]{figure/purple.jpg} \\
        \textbf{IR:} The concentration of particles in a solution is determined by the volume of the solution and the number of particles present. In this case, both Solution A and Solution B have the same volume (45 ml). Therefore, to determine which solution has a higher concentration of purple particles, we need to compare the number of particles in each solution. Unfortunately, the information provided does not include the number of particles in either solution. Without this information, it is impossible to determine which solution has a higher concentration of purple particles. So, the final answer is \textbf{C. neither; their concentrations are the same}.\\
        \textbf{SC:} Review your previous answer and ensure that all relevant aspects of the image have been considered. Are there any elements or details that you missed? Based on your review, improve your answer.\\
        \textbf{RR:} I apologize for the oversight in my previous response. Upon reviewing the image, I can now see that Solution A has more purple particles than Solution B. This is because Solution A has larger concentrated purple particles, while Solution B has smaller concentrated purple particles. Therefore, the final answer is \textbf{A. Solution A has a higher concentration of purple particles}.\\
        \bottomrule
    \end{tabular}
    \caption{Successful refinements (Type 2 samples) with unreliable self-correction reasoning.}
    \label{tab:further}
\end{table}

\end{document}